%% file: main.tex
\documentclass[runningheads]{llncs}

\usepackage{eccv}

\usepackage{eccvabbrv}

\usepackage{graphicx}
\usepackage{logo}
\usepackage{booktabs}

\usepackage{multirow}
\usepackage{marvosym}

\usepackage[accsupp]{
  axessibility
} 

\usepackage{hyperref}
\hypersetup{colorlinks=true, allcolors=eccvblue}

\usepackage{orcidlink}

\definecolor{DColor}{RGB}{255,60,0}

\begin{document}

  \title{TabletopGen: Tabletop Scene Generation and Interactive Simulation for Robotic Manipulation}

  \titlerunning{TabletopGen}

  \author{
  Ziqian Wang\inst{1,2,3} \and
  Yonghao He\inst{3}\textsuperscript{\dag} \and
  Licheng Yang\inst{1,2} \and
  Wei Zou\inst{1,2} \and
  Hongxuan Ma\inst{2} \and
  Liu Liu\inst{4} \and
  Wei Sui\inst{3} \and
  Yuxin Guo\inst{1,2} \and
  Hu Su\inst{2}\textsuperscript{\Letter}
  }

  \authorrunning{Z.~Wang et al.}
  
  \institute{
  School of Artificial Intelligence, University of Chinese Academy of Sciences, Beijing, China
  \and
  State Key Laboratory of Multimodal Artificial Intelligence Systems (MAIS), Institute of Automation, Chinese Academy of Sciences, Beijing, China
  \and
  D-Robotics, Beijing, China
  \and
  Horizon Robotics, Beijing, China\\
  \email{ \{wangziqian2024,hu.su\}@ia.ac.cn}\\
  \textsuperscript{\dag}Project Leader \quad \textsuperscript{\Letter}Corresponding Author\\
  \url{https://d-robotics-ai-lab.github.io/TabletopGen.project/}
  }

  \setcompanylogo{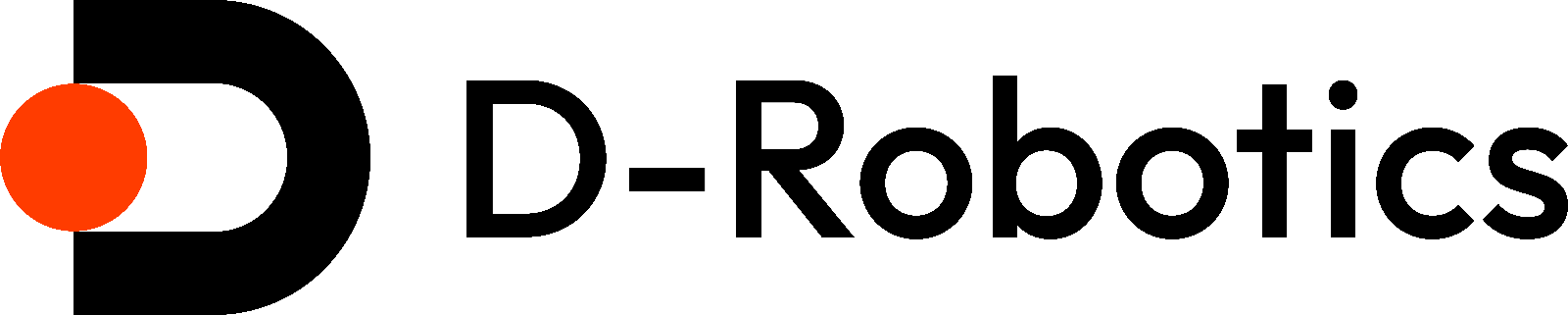}
  \setpartnerlogoA{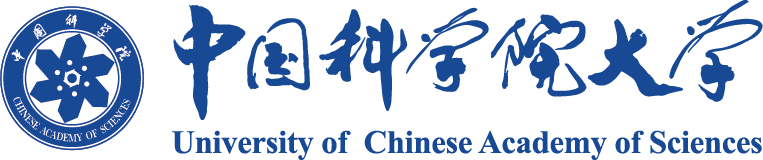}
  \setpartnerlogoB{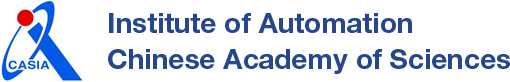}
  \setpartnerlogoC{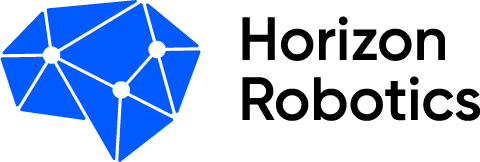}

  \makeatletter
  \def\@maketitle{%
   \newpage
   \markboth{}{}%
   \vspace*{-0.75cm}%
   \makelogos
   {\color{DColor}\noindent\rule{\textwidth}{0.8pt}}\vskip .45cm
   \def\lastand{\ifnum\value{@inst}=2\relax
                   \unskip{} \andname\
                \else
                   \unskip \lastandname\
                \fi}%
   \def\and{\stepcounter{@auth}\relax
            \ifnum\value{@auth}=\value{@inst}%
               \lastand
            \else
               \unskip,
            \fi}%
   \begin{center}%
   \let\newline\\
   {\Large \bfseries\boldmath
    \pretolerance=10000
    \@title \par}\vskip .8cm
  \if!\@subtitle!\else {\large \bfseries\boldmath
    \vskip -.65cm
    \pretolerance=10000
    \@subtitle \par}\vskip .8cm\fi
   \setbox0=\vbox{\setcounter{@auth}{1}\def\and{\stepcounter{@auth}}%
   \def\thanks##1{}\@author}%
   \global\value{@inst}=\value{@auth}%
   \global\value{auco}=\value{@auth}%
   \setcounter{@auth}{1}%
  {\lineskip .5em
  \noindent\ignorespaces
  \@author\vskip.35cm}
   {\small\institutename}
   \end{center}%
  }
  \makeatother

  \maketitle
  
  \input{sec/0_abstract}

  \begin{figure}[t]
    \centering
    \includegraphics[width=\textwidth]{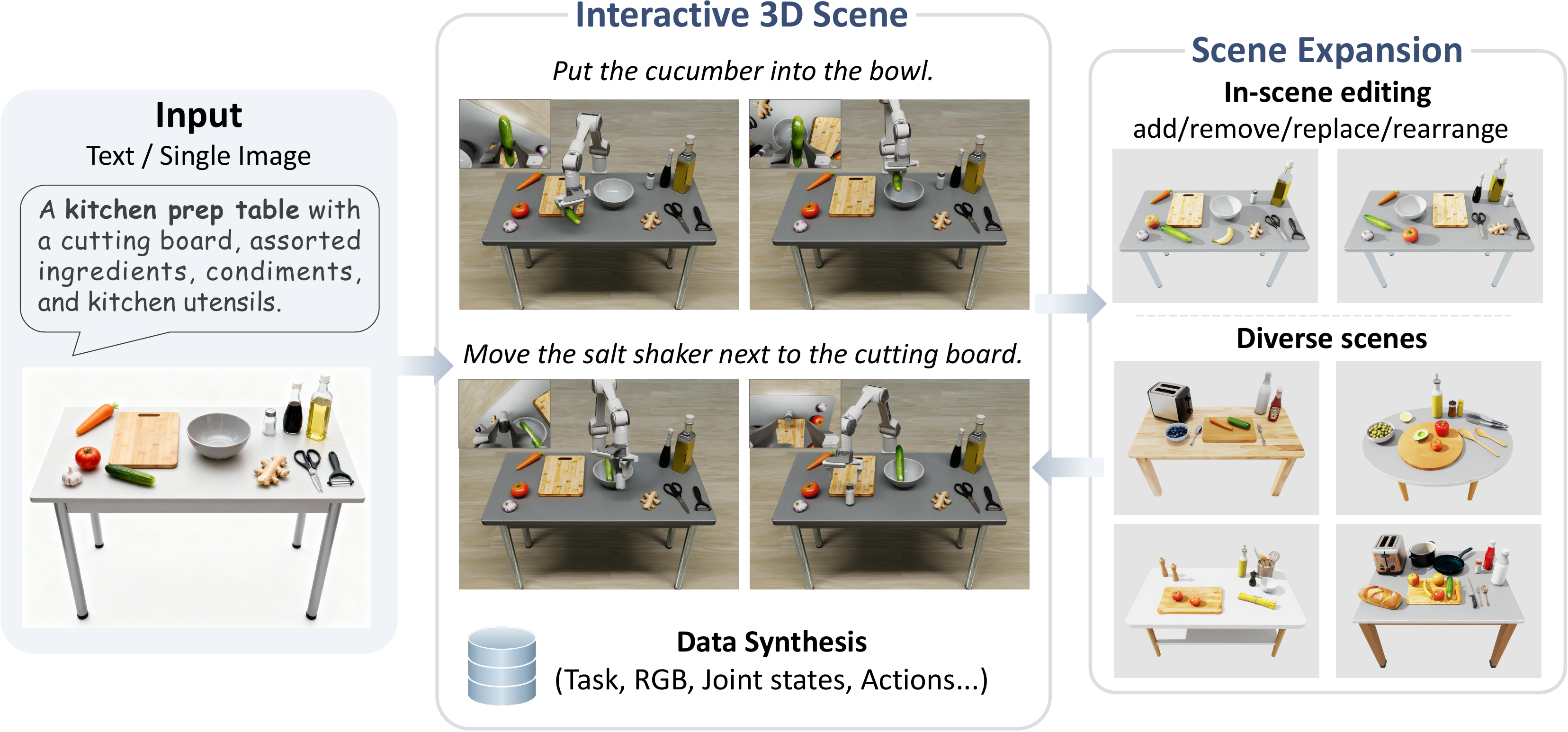}
    \caption{We present \textbf{TabletopGen}, a training-free, fully automatic framework for generating tabletop simulation scenes. Given a text description or a single image, TabletopGen produces high-fidelity, detail-rich, instance-level interactive tabletop environments that can be directly used for robotic manipulation tasks. It supports multimodal data synthesis within the simulation and enables scene expansion---either via in-scene editing or by generating diverse scenes under the same description---thereby providing a richer coverage of simulated environments for large-scale data synthesis. We showcase more tabletop scenes of various types and styles in the supplementary.}
    \label{fig:teaser}
  \end{figure}

  \input{sec/1_introduction}
  \input{sec/2_relatedwork}
  \input{sec/3_method}
  \input{sec/4_experiments}
  \input{sec/5_conclusion}

  \bibliographystyle{splncs04}
  \bibliography{main}

  \input{sec/supp}

\end{document}

%% file: sec/0_abstract.tex
\begin{abstract}

Simulation provides a low-cost, scalable pathway to large-scale robotic manipulation data collection. However, existing 3D scene generation methods can rarely be applied directly to manipulation data synthesis, as their generated scenes often lack instance-level interactivity and physical plausibility. Focusing on tabletop manipulation, we propose \textbf{TabletopGen}, a training-free and automated tabletop scene generation and interactive simulation engine. Starting from text or a single image, we first obtain independent 3D object models via generative instance extraction. Second, we introduce a novel pose and scale alignment approach that recovers a collision-free scene layout using a Differentiable Rotation Optimizer and a Top-View Spatial Alignment mechanism. Finally, we assemble the generated scene in a physics simulator with collision geometry, yielding a stable, interactable environment for synthesizing multimodal manipulation data. Extensive experiments and user studies demonstrate that TabletopGen achieves state-of-the-art performance in visual fidelity, layout accuracy, and physical plausibility. Furthermore, we validate the executability of the collected trajectories on a real robotic arm via zero-shot real-to-sim-to-real policy transfer, indicating that TabletopGen can serve as a reliable data engine for robotic manipulation data synthesis.
\keywords{3D Scene Generation \and Tabletop Environments \and Interactive Simulation \and Manipulation Data Synthesis}

\end{abstract}

%% file: sec/1_introduction.tex
\section{Introduction}
\label{sec:introduction}

The tabletop environment is the "last meter" of robotic manipulation: it serves as the fundamental platform for most fine-grained interaction tasks such as grasping, placing, and assembly, and acts as a touchstone for evaluating robots' capabilities in complex manipulation \cite{tang2023icra:rgbonly,hao2025mesatask,zhang2023lohoravenslonghorizonlanguageconditionedbenchmark}. With the rapid development of Embodied AI and Vision-Language-Action (VLA) models, learning manipulation policies increasingly relies on massive, highly generalizable interaction data \cite{open_x_embodiment_rt_x_2023,black2024pi_0,bu2025agibot,intelligence2025pi_}. However, collecting data in real-world tabletop settings is often costly, inefficient, and difficult to scale up for broad coverage and generalization. In contrast, synthesizing data in simulation offers a highly promising route to obtain diverse interaction demonstrations in a low-cost and controllable manner\cite{chen2025robotwin,tian2025interndata,yu2025metascenes}. Therefore, automatically generating high-fidelity, instance-interactive, physically plausible 3D tabletop scenes---and efficiently synthesizing manipulation data therein for planning and learning---is a critical step toward improving the generalization of robotic policies.

However, existing 3D scene generation methods often struggle to directly produce simulation-ready tabletop environments. Retrieval-based \cite{dai2024acdc,yang2024holodeck,ccelen2024design,hao2025mesatask} approaches acquire instance models from fixed asset libraries \cite{li2024behavior1khumancenteredembodiedai,deitke2023objaverse,NEURIPS2023_70364304,fu20213d}, often causing insufficient diversity and geometric mismatch. Large language model (LLM)-based text-driven methods \cite{yang2024holodeck,ccelen2024design,gu2025artiscene,fu2024anyhome,sun2025layoutvlm,aguina2024open} mostly focus on room-scale layouts, failing to handle the high-density, small-object functional arrangements and strong semantic constraints inherent to tabletop scenes. While single-image reconstruction methods \cite{huang2025midi,li2024dreamscene,chung2023luciddreamer,ardelean2025gen3dsr,meng2025scenegen} can generate visually similar scenes, they are limited by incomplete observation from a single viewpoint and the lack of physical priors (\eg, treating the tabletop as a Z-support and XY-boundary). As a result, they often produce instance meshes with fused or broken topology, as well as physically implausible layouts such as interpenetration and floating objects. Consequently, these scenes cannot be reliably imported into simulators for motion planning, as low-quality instances and physically invalid layouts compromise collision modeling and reachability, making it difficult to automatically generate large-scale, executable manipulation data.

To break this data bottleneck, we propose \textbf{TabletopGen}, a tabletop environment generation engine. As illustrated in \cref{fig:teaser}, TabletopGen takes text or a single image as input and automatically produces diverse, instance-level, physically plausible tabletop scenes that can be directly used for motion planning and data synthesis. Our framework decouples the complex scene construction problem into three progressive stages. First, in the generative instance extraction stage, we extract and complete high-quality 2D instances from text-to-image synthesized or real-world images, and reconstruct them into 3D assets under a unified canonical coordinate system, thereby removing the reliance on fixed libraries and ensuring geometric completeness of instances. Next, to accurately estimate the rotation, translation, and scale of instances, we propose \textbf{a novel two-phase pose and scale alignment approach}: we use a Differentiable Rotation Optimizer (DRO) to estimate rotations, and introduce a Top-View Spatial Alignment (TSA) mechanism to provide physical priors and infer globally consistent translations and scales, yielding physically plausible layouts. Finally, we assemble the instances in a simulator and attach physical properties, thereby producing realistic and interactive 3D tabletop scenes that conform to real-world logic. Moreover, TabletopGen supports fast scene derivation by adding, removing, replacing, and rearranging objects, and can generate diverse scenes under the same text description. This enables low-cost scene scaling and broader simulation environment coverage for large-scale data synthesis.

Thanks to instance independence and physically plausible layouts, the simulation scenes generated by TabletopGen can directly support data synthesis for manipulation policy learning. To validate the real-world utility of the synthesized demonstrations, we further conduct a zero-shot real-to-sim-to-real policy transfer experiment. Given a single photograph of a real tabletop, our framework reconstructs an instance-level simulated scene, synthesizes manipulation trajectories with domain randomization, and uses them to fine-tune a robot policy. The resulting policy is then directly deployed on a real robotic arm without any real-world data fine-tuning. Successful real-world executions across multiple tasks indicate that the generated assets, spatial layouts, and physical properties are sufficiently accurate to support downstream sim-to-real policy learning, validating TabletopGen as a practical tabletop manipulation data engine.

In summary, our main contributions are as follows:
\begin{itemize}
    \item We present TabletopGen, an automated, training-free 3D tabletop scene generation framework. Starting from text or a single image, it generates simulation scenes that jointly achieve visual realism, instance-level interactivity, and physical plausibility, providing an efficient data synthesis route for policy learning in robotic manipulation tasks.
    
    \item We propose a novel pose and scale alignment method tailored for tabletop scenes. Via the decoupled DRO and TSA phases, we recover visually consistent and physically plausible tabletop layouts from the 2D reference.
    
    \item We conduct comprehensive system-level validation. Extensive experiments show that TabletopGen significantly outperforms existing methods in scene quality and physical validity. Furthermore, our zero-shot real-to-sim-to-real policy transfer experiments confirm the reliability of the generated scenes and synthetic trajectories for real-world robotic manipulation policy learning.
\end{itemize}

%% file: sec/2_relatedwork.tex
\section{Related Works}
\label{sec:relatedworks}

Large-scale, high-quality robot data is crucial for embodied AI. For example, RT-1 \cite{brohan2022rt} used approximately 130k trajectories collected by 13 robots over 17 months, and Octo \cite{team2024octo} was trained on around 800k trajectories. The data for these studies are generally collected manually, which is costly, inefficient, and difficult to scale and generalize. With the advancement of physics simulation engines such as Isaac Sim \cite{NVIDIA_Isaac_Sim} and SAPIEN \cite{xiang2020sapien}, efficiently synthesizing data in simulation offers a promising solution to this challenge \cite{zhao2020sim}. Meeting the demand for high-quality data generation requires building simulation scenes that are diverse and highly realistic. Related research primarily follows two directions: digital-twin reconstruction from real observations, and generative 3D scene synthesis.

\subsection{Digital Twin and Real-to-Sim-to-Real}
 These methods \cite{torne2024rialto,han2025re3sim,torne2024casher,li2025rose} reconstruct digital-twin scenes from real-world videos or image sequences, and collect manipulation demonstrations or optimize policies within them. By providing spatial geometry consistent with the real world, they can support high-precision Real-to-Sim-to-Real closed-loop deployment. However, their scene construction often relies on multi-view scanning or manual intervention. This strong dependence on real-world observations limits the diversity of simulated scenes, making it difficult to satisfy the need for massive scenes in embodied learning.

\subsection{Generative 3D Scene Synthesis}
To overcome the diversity bottleneck of real-scene inputs, an increasing number of works have begun exploring automatic 3D scene generation from text or single image, attempting to serve them as simulation environments for embodied AI.

\textbf{Text-driven methods} use LLMs for semantic and spatial reasoning to either directly generate 3D layouts \cite{NEURIPS2023_3a7f9e48,yang2024llplace3dindoorscene,ocal2024sceneteller}, or first produce scene graphs or spatial constraints and then optimize layouts via solvers \cite{sun2025layoutvlm,yang2024holodeck,fu2024anyhome,ccelen2024design,aguina2024open,gao2024graphdreamer,li2024dreamscene,ling2025scenethesislanguagevisionagentic}. These methods are typically designed for room-scale scenes and can support tasks like navigation and planning. However, they struggle to capture the high-density, semantics-constrained functional layouts of small objects that are unique to tabletops, and thus are not well-suited for fine-grained robotic manipulation. Moreover, most methods rely on retrieving 3D assets from fixed libraries, which limits style consistency and asset flexibility. Recently, MesaTask \cite{hao2025mesatask} proposed a text-driven scene generation method targeting tabletop tasks, but it also relies on an asset library and places objects only on a rectangular plane, without modeling the full 3D structure of the table.

\textbf{Image-driven methods} aim to recover visually consistent 3D scenes from a single image. Retrieval-based approaches \cite{dai2024acdc,gao2024diffcad,gumeli2022roca, zook2025grs} are constrained by limited asset libraries and thus cannot precisely reconstruct target objects. Diffusion-based holistic generation methods \cite{huang2025midi,chen2025housecrafter,yu2025wonderworld} can end-to-end generate scenes consistent with the input view. Yet, due to incomplete single-view observations, they often produce fused or broken instance meshes. In addition, lacking dedicated tabletop datasets and physical priors, their generated scenes frequently exhibit physically implausible layouts such as interpenetration, floating objects, or objects falling outside the tabletop boundary. Compositional methods \cite{ardelean2025gen3dsr,gu2025artiscene,meng2025scenegen,han2025reparo,yao2025cast,hu2025flashsculptormodular3d} improve flexibility by per-instance generation and alignment; however, they typically rely on monocular depth back-projection for pose estimation. In dense tabletop layouts of small objects,  camera calibration or depth estimation errors can be amplified and accumulated in scale and pose computation, and these methods also lack physical constraints during scene assembly, hindering globally consistent and physically feasible tabletop layouts.

Unlike the above methods, TabletopGen is a simulation data engine specifically designed for tabletop robotic manipulation. It ensures high-fidelity, instance-level 3D reconstruction from a single image, while leveraging the diversity of text-to-image generation to expand scenes infinitely. More importantly, we propose a novel decoupled pose and scale alignment approach: DRO precisely optimizes instance rotations using visual features, and TSA introduces physical priors via top-view and stacking relations to constrain scene boundaries and support relations. As a result, TabletopGen generates globally consistent and physically plausible tabletop simulation environments that can be directly used for manipulation data synthesis and zero-shot real-to-sim-to-real transfer.

%% file: sec/3_method.tex
\section{Method}
\label{sec:method}

\begin{figure}[t]
   \centering
   \includegraphics[width=0.98\linewidth]{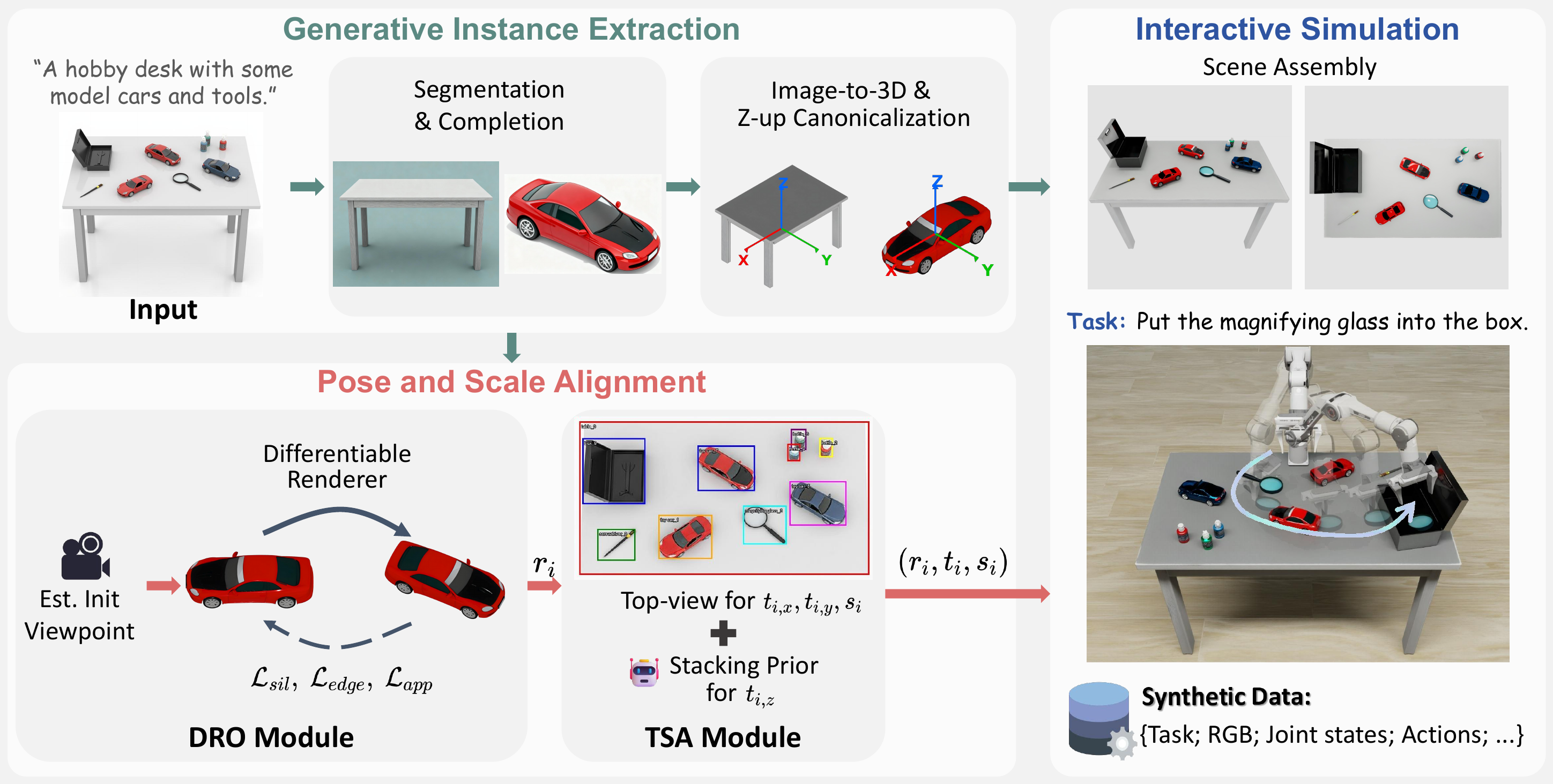}

   \caption{\textbf{Overview of our TabletopGen Framework.} Given a text prompt (first converted into a reference image via Text-to-Image) or a single image, we proceed in three stages: (1) \textbf{Generative Instance Extraction} segments and completes 2D instances, and reconstructs Z-up canonicalized, directly placeable 3D assets. (2)\textbf{Pose and Scale Alignment} recovers a physically feasible layout: The \textbf{DRO (Differentiable Rotation Optimizer)} optimizes rotation with a tri-modal loss, while the \textbf{TSA (Top-view Spatial Alignment)} infers translation and scale using a synthesized top-view and stacking priors, together with an anchor object selected by RMA-Score. (3) \textbf{Interactive Simulation} assembles the scene in a physics simulator, performs motion planning for manipulation tasks and executes collision-free trajectories, thereby synthesizing multimodal interaction data.}
   \label{fig:overview}
\end{figure}

Given a text description $T$ or an image $I$, our goal is to generate a tabletop scene $S = \{o_{i}\}_{i=1}^{n}$ that is interactive in a physics simulator. Each instance $o_{i}$ is defined by its 3D model $m_{i}$, scale $s_{i} \in \mathbb{R}^{3}$, translation $t_{i} \in \mathbb{R}^{3}$, and rotation $r_{i} \in \mathbb{R}$ around the vertical axis. As shown in \cref{fig:overview}, we adopt a multi-stage architecture: we first generate canonicalized instance assets, then obtain a globally consistent layout via a decoupled pose and scale alignment approach, and finally assemble the scene in a simulator to support robotic manipulation data synthesis. Following prior works \cite{gu2025artiscene,zhou2024layoutyour3d}, we argue that images provide more realistic and precise layout information than text. Thus, for a text input $T$, an LLM expands it into a detailed prompt that specifies the scene style, object set, and their functional layout, and a text-to-image model synthesizes a visually realistic and plausibly arranged reference image $I_{ref}$; image inputs serve directly as $I_{ref}$.

\subsection{Generative Instance Extraction}
\label{sec:instance extraction}

To enable instance-level interaction in simulation, we first extract and reconstruct geometrically complete and coordinate-canonical 3D assets from $I_{ref}$. We begin by using a multimodal large language model (MLLM) for open-vocabulary category recognition over tabletop scene objects (including the table itself), and combine it with GroundedSAM-v2 \cite{ren2024grounded} to obtain instance masks. Because severe single-view occlusions often yield masks with holes or blurred boundaries, we discard traditional inpainting and instead utilize a multimodal generative model to complete and redraw each instance mask. The completed instance images are used to generate high-quality 3D meshes $m_{i}^{\prime}$ via Image-to-3D. However, the local coordinate system of $m_{i}^{\prime}$ is typically arbitrary, and directly using it can lead to unplaceable states such as being tilted or inverted. Therefore, guided by the visual cues from $I_{ref}$ and the semantic priors from the MLLM, we apply rotation correction to align the local vertical axis of each $m_{i}^{\prime}$ to the world Z-up direction, yielding canonical assets $m_{i}$ that can be directly placed in tabletop scenes.

\subsection{Pose and Scale Alignment}
\label{sec:pose}

The second and most critical stage is to estimate each instance's \textbf{rotation}
$r_{i}$, \textbf{translation} $t_{i}$, and \textbf{scale} $s_{i}$, so that all canonical 3D models $m_{i}$ can be assembled into a physically plausible scene consistent with $I_{ref}$. We propose a novel approach that decouples this problem into two phases: an instance's rotation around the vertical axis typically does not change the overall layout but can substantially alter its silhouette and texture distribution, impacting the visual consistency with the input. Therefore, our DRO phase precisely optimizes the rotation based on visual information. In contrast, translation and scale directly affect the spatial relations. Our TSA phase utilizes a top-view image to compensate for single-view observation and, combined with the constraints of physical priors, uniformly infers globally consistent translations and scales, thereby obtaining an executable physical layout.

\subsubsection{Differentiable Rotation Optimizer (DRO).}
In this phase, we estimate $r_{i}$ for each instance. Under the initial camera viewpoint (azimuth, elevation) estimated by MLLM, we render each $m_{i}$ using a differentiable renderer and obtain a textured image $I_{render}(r_{i})$, a soft silhouette $\hat{S}(r_{i})$, and a soft edge map $\hat{E}(r_{i})$ derived from $\hat{S}(r_{i})$ using a Sobel filter. We then define a
\textbf{tri-modal matching loss} $\mathcal{L}_{rot}$ to align the render with the
target instance $I_{instance}$ and its mask $S$:
\begin{equation}
  \mathcal{L}_{rot}(r_{i}) = \lambda_{s}\mathcal{L}_{sil}(r_{i}) + \lambda_{e}\mathcal{L}
  _{edge}(r_{i}) + \lambda_{a}\mathcal{L}_{app}(r_{i}) \label{eq:Lrot}
\end{equation}
where $\lambda_{s}$, $\lambda_{e}$ and $\lambda_{a}$ are weighting coefficients.
The first term, $\mathcal{L}_{sil}$, is a soft IoU loss for shape consistency:
\begin{equation}
  \mathcal{L}_{sil}(r_{i}) = 1 - \frac{\sum(\hat{S}(r_{i}) \cdot S)}{\sum(S +
  \hat{S}(r_{i}) - S \cdot \hat{S}(r_{i}))}\label{eq:Lsil}
\end{equation}

The second, $\mathcal{L}_{edge}(r_{i})$, is a one-sided Chamfer loss that
matches contours. It penalizes the distance from $\hat{E}(r_{i})$ to the nearest
ground-truth edge. A distance transform $D_{S}$ is pre-computed from Canny edges
of the target mask $S$. The loss is the weighted average distance over all pixels
$x$:
\begin{equation}
  \mathcal{L}_{edge}(r_{i}) = \frac{\sum_{x}D_{S}(x) \cdot \hat{E}(x,r_{i})}{\sum_{x}\hat{E}(x,r_{i})}
  \label{eq:Ledge}
\end{equation}

Finally, $\mathcal{L}_{app}$ is a perceptual feature loss that ensures
appearance similarity with DINOv2~\cite{oquab2023dinov2} features $\Phi$:
\begin{equation}
  \mathcal{L}_{app}(r_{i}) = || \Phi(I_{render}(r_{i})) - \Phi(I_{instance}) ||_{2}
  ^{2} \label{eq:Lapp}
\end{equation}

We find the optimal rotation $\hat{r}_{i}$ by minimizing this loss via gradient
descent: $\hat{r}_{i} = \arg\min_{r_i}\mathcal{L}_{rot}(r_{i})$. This process
can also optionally refine the camera pose.

\subsubsection{Top-View Spatial Alignment (TSA).}
In this phase, we incorporate two types of physical priors to achieve a physically plausible layout estimation: (1) we employ an MLLM to infer the stacking relations among instances for placement, thereby avoiding floating and vertical interpenetration; (2) we utilize a multimodal generative model to synthesize a top-view of the scene, $I_{top}$, and extract the 2D bounding boxes of all instances. This top-view prior naturally ensures that all objects are constrained within the valid tabletop boundaries.

Building upon this, we estimate the translation and scale. For each instance, let $r_{img}$ and $A_{px}$ denote the pixel aspect ratio and pixel area of its top-view bounding box, respectively. We query the MLLM for its commonsense physical size and apply the estimated $r_i$ to obtain its physical aspect ratio $r_{phys}$ projected onto the xy-plane. We then propose a Ratio-Matched and Area-Weighted Score (RMA-Score) to select a reliable scaling anchor:
\begin{equation}
    \varepsilon_{ratio} = |\log r_{phys} - \log r_{img}|, \quad \text{RMA}(i) = \frac{A_{px}(i)}{1 + (\varepsilon_{ratio}/\tau)^2}
\end{equation}
where $\tau$ is a tolerance hyperparameter. This score favors objects with large areas (for stability) and low aspect-ratio mismatch (to ensure geometric consistency). The instance with the highest RMA-Score is selected as the anchor to compute a global scaling factor $\alpha$ (meters/pixel), thereby precisely inferring the 3D scale $s_i$ and the horizontal translation coordinates $(t_{i,x}, t_{i,y})$ for all instances. The vertical translation $t_{i,z}$ is then determined based on the stacking relations, ultimately yielding a physically plausible 3D layout $\{r_i, t_i, s_i\}$.

\subsection{Interactive Simulation}

We import each $m_i$ into a physics simulator, Isaac Sim, and apply the corresponding $\{r_i, t_i, s_i\}$ transforms. To support realistic rigid-body dynamic interactions, we generate collision geometries for each instance via convex decomposition and enable gravity and friction parameters, thereby turning the scene into an interactive simulated tabletop environment. Furthermore, by adding, removing, replacing, or locally rearranging instances, we can expand the original scene into a massive number of variants for large-scale data synthesis.

Within the simulation environment, we support both manually specified manipulation tasks and the automatic generation of reasonable interaction tasks using an LLM based on the instances' semantic labels and physical properties. Given a task, we employ a motion planner \cite{sundaralingam2023curobo} to automatically compute collision-free trajectories. During trajectory execution by the robotic arm, we can collect rich multimodal interaction data at low cost, including multi-view RGB images, joint states, actions, and more. Through this complete closed loop---from automated scene generation to automated data collection---TabletopGen provides a continuous stream of high-quality synthetic data for training VLA models with strong generalization capabilities.

%% file: sec/4_experiments.tex
\section{Experiments}
\label{sec:experiments}

\subsection{Setup}

\subsubsection{Implementation details.}
We use ChatGPT \cite{achiam2023gpt} for vision-language reasoning, Seedream \cite{seedream2025seedream} for image synthesis, and Hunyuan3D-3.0 \cite{lai2025hunyuan3d25highfidelity3d} to generate 3D meshes. 
In DRO, we render with PyTorch3D \cite{ravi2020pytorch3d} and run a coarse search over the [0$^\circ$, 360$^\circ$) range at 5$^\circ$ step size to select the 8 candidate angles with the lowest loss at first. Each candidate is then refined using the Adam optimizer \cite{kingma2014adam} (lr $=3 \times 10^{-2}$, 140 steps; $\lambda_{s}=0.5$, $\lambda_{e}=0.5$, $\lambda_{a}=2.0$). In TSA, we set the RMA-Score tolerance to $\tau=0.25$. The prompts and step-wise intermediate outputs are provided in the supplementary materials.

\subsubsection{Baselines.}
Since our tabletop scenes are primarily generated from 2D reference, we compare against representative single-image 3D scene generation methods: ACDC \cite{dai2024acdc}, Gen3DSR \cite{ardelean2025gen3dsr}, and MIDI \cite{huang2025midi}. To illustrate our text-to-scene capability, we also compare with MesaTask~\cite{hao2025mesatask} and Holodeck-table, a tabletop-adapted version of Holodeck~\cite{yang2024holodeck} following the adaptation strategy used in MesaTask.

\subsubsection{Metrics.}
Following prior works \cite{gu2025artiscene,ling2025scenethesislanguagevisionagentic,meng2025scenegen,hao2025mesatask,yao2025cast,han2025reparo,hu2025flashsculptormodular3d}, we comprehensively evaluate both visual consistency and simulation usability: (1) LPIPS \cite{zhang2018unreasonable}, DINOv2 \cite{oquab2023dinov2}, and CLIP \cite{radford2021learning} for perceptual similarity, visual consistency, and semantic alignment; (2) Physical plausibility via the proportion of colliding object pairs (Col\_O) and the percentage of scenes containing collisions (Col\_S); (3) Multi-dimensional GPT Evaluation using GPT-4o \cite{hurst2024gpt} to rate Visual Fidelity (VF), Image Alignment (IA) and Physical Plausibility (PP) on a 1-7 scale, and directly provide an Overall Ranking (OR) across methods.

For text-to-scene comparisons, where no unique reference image is available, we evaluate the generated scenes with GPT-4o under the same input text prompts. Specifically, we rate Consistency with Text (CwT), Object Quality and Recognizability (OQR), Layout Coherence and Realism (LCR), and Physical Plausibility (PP) on a 1-7 scale, and further report Col\_O and Col\_S.

\subsection{Comparisons}

We construct a diverse test set containing 78 samples to evaluate the robustness and generalization capability of our method. The test set is built from both synthesized images and real-world photographs, and spans multiple table shapes---square, round, and triangular. For functional categories, we follow the taxonomy used in MesaTask \cite{hao2025mesatask} and recommendations from ChatGPT, covering office tables, dining tables, workbenches, and crafting tables. By comparing TabletopGen with different baseline methods under identical inputs, we demonstrate its significant superiority in generating diverse tabletop scenes.

\begin{table}[t]
  \caption{\textbf{Quantitative comparison with image-driven methods.} Our method TabletopGen achieves the best overall performance across all metrics, showing substantially superior results in Visual \& Perceptual quality, GPT-4o Evaluation, and near-zero Collision Rates compared to all baselines.}
  \label{tab:quantitative}
  \centering
    \begin{tabular}{@{}lcccccccccc@{}}
      \toprule
      \multirow{2}{*}{\textbf{Method}} & \multicolumn{3}{c}{\textbf{Visual \& Perceptual}} & \multicolumn{5}{c}{\textbf{GPT Evaluation}} & \multicolumn{2}{c}{\textbf{Collision Rate(\%)}} \\
      \cmidrule(lr){2-4} \cmidrule(lr){5-9} \cmidrule(lr){10-11}
      & LPIPS$\downarrow$ & DINOv2$\uparrow$ & CLIP$\uparrow$ & VF$\uparrow$ & IA$\uparrow$ & PP$\uparrow$ & Avg.$\uparrow$ & OR$\downarrow$ & Col\_O$\downarrow$ & Col\_S$\downarrow$ \\
      \midrule
      ACDC    & 0.5124 & 0.3775 & 0.6696 & 2.38 & 1.90 & 2.57 & 2.28 & 3.55 & 8.23  & 67.95 \\
      Gen3DSR & 0.4891 & 0.5602 & 0.8636 & 2.92 & 4.17 & 3.32 & 3.47 & 3.02 & 16.88 & 85.90 \\
      MIDI    & 0.4559 & 0.7070 & 0.8867 & 4.22 & 4.48 & 4.30 & 4.33 & 2.32 & 17.39 & 98.72 \\
      Ours    & \textbf{0.4483} & \textbf{0.8383} & \textbf{0.9077} & \textbf{6.06} & \textbf{6.30} & \textbf{6.22} & \textbf{6.19} & \textbf{1.08} & \textbf{0.42} & \textbf{7.69} \\
      \bottomrule
    \end{tabular}
\end{table}

\begin{table}[t]
\centering
\caption{\textbf{Quantitative comparison with text-driven methods.} Our method achieves substantially better text consistency, object quality, layout realism, and overall scene quality, while maintaining near-zero collision rates.}
\label{tab:text_comparison}
\begin{tabular}{@{}lccccccc@{}}
\toprule
\multirow{2}{*}{\textbf{Method}} &
\multicolumn{5}{c}{\textbf{GPT Evaluation}} &
\multicolumn{2}{c}{\textbf{Collision Rate(\%)}} \\
\cmidrule(lr){2-6} \cmidrule(lr){7-8}
& CwT$\uparrow$ & OQR$\uparrow$ & LCR$\uparrow$ & PP$\uparrow$ & Avg.$\uparrow$
& Col\_O$\downarrow$ & Col\_S$\downarrow$ \\
\midrule
Holodeck-table & 3.57 & 4.17 & 3.90 & 4.70 & 4.08 & \textbf{0.00} & \textbf{0.00} \\
MesaTask       & 3.63 & 4.23 & 4.07 & 4.57 & 4.12 & 4.95 & 63.33 \\
Ours           & \textbf{6.63} & \textbf{6.10} & \textbf{6.07} & \textbf{6.27} & \textbf{6.27} & 0.20 & 6.67 \\
\bottomrule
\end{tabular}
\end{table}

\subsubsection{Quantitative Evaluation.}
\cref{tab:quantitative} reports that our method achieves the best performance across all metrics in the three evaluation aspects, clearly outperforming all image-driven baselines. For \textbf{Visual \& Perceptual quality}, TabletopGen attains the best LPIPS, DINOv2, and CLIP scores. We attribute this to our "instance-first, then alignment" design, which first ensures high-fidelity per-instance reconstruction and subsequently achieves precise pose and scale alignment. Under the \textbf{GPT Evaluation}, TabletopGen achieves the highest average score (6.19)---a 43\% improvement over the second-best MIDI and the top Overall Ranking (OR). This demonstrates that our results are perceived as more realistic and semantically plausible. The \textbf{Collision Rate} shows the most significant performance gap. As previously analyzed, existing 3D scene generation methods struggle to capture the high-density tabletop layouts with complex spatial relations. Their Col\_O typically lies around 8\%--17\%, and the scene-level Col\_S reaches 67\%--98\%. In contrast, TabletopGen precisely recovers layouts via its DRO and TSA stages. Our Col\_O is near zero (0.42\%), and Col\_S remains only 7.69\%. This overwhelming advantage demonstrates our method's robustness and practical value in producing collision-free, interactive tabletop scenes.

For text-to-scene generation, we evaluate 30 text-scene pairs covering diverse tabletop types. As shown in \cref{tab:text_comparison}, TabletopGen achieves clearly higher GPT-4o scores than MesaTask and Holodeck-table, improving the average rating from 4.12/4.08 to 6.27. Holodeck-table obtains zero collisions mainly through hard constraints on 2D bounding boxes, which often leads to sparse and less realistic layouts, while MesaTask still suffers from frequent scene-level collisions. In contrast, TabletopGen maintains near-zero mesh-level collisions while preserving realistic and task-consistent tabletop arrangements.

\subsubsection{User Study.}

We further conducted a comprehensive user study to assess perceptual quality and human preference, involving 128 participants. Each participant was randomly assigned 8 scenes; to ensure fairness, the display order of results from different methods was fully randomized. Participants rated the generated scenes on three
criteria---Visual Fidelity (VF), Image Alignment (IA), and Physical Plausibility (PP)---using a 7-point scale. As shown in \cref{tab:user}, our method significantly outperforms all baselines on every criterion. Our average score of 5.56 was substantially higher than the second-best method (3.57). Most importantly, when asked for an Overall Preference (OP), our results were selected in 83.13\% of cases, confirming that our generated scenes are perceived as more realistic and physically plausible by human evaluators.

\begin{table}
  \caption{\textbf{User study results.} Our method TabletopGen attains the highest mean scores on Visual Fidelity (VF), Image Alignment (IA), and Physical Plausibility (PP), and the best overall preference (OP), being selected in 83.13\% of cases.}
  \label{tab:user}
  \setlength{\tabcolsep}{7pt}
  \centering
  \begin{tabular}{lccccc}
    \toprule \textbf{Method} & \textbf{VF$\uparrow$} & \textbf{IA$\uparrow$} & \textbf{PP$\uparrow$} & \textbf{Avg.$\uparrow$} & \textbf{OP(\%)} \\
    \midrule ACDC            & 3.00                  & 2.46                  & 2.87                  & 2.78                    & 2.38            \\
    Gen3DSR                  & 2.95                  & 3.25                  & 3.19                  & 3.13                    & 2.88            \\
    MIDI                     & 3.82                  & 3.57                  & 3.32                  & 3.57                    & 11.61           \\
    Ours                     & \textbf{5.62}         & \textbf{5.50}         & \textbf{5.56}         & \textbf{5.56}           & \textbf{83.13}  \\
    \bottomrule
  \end{tabular}
\end{table}

\begin{figure}[t]
  \centering
  \includegraphics[width=1\linewidth]{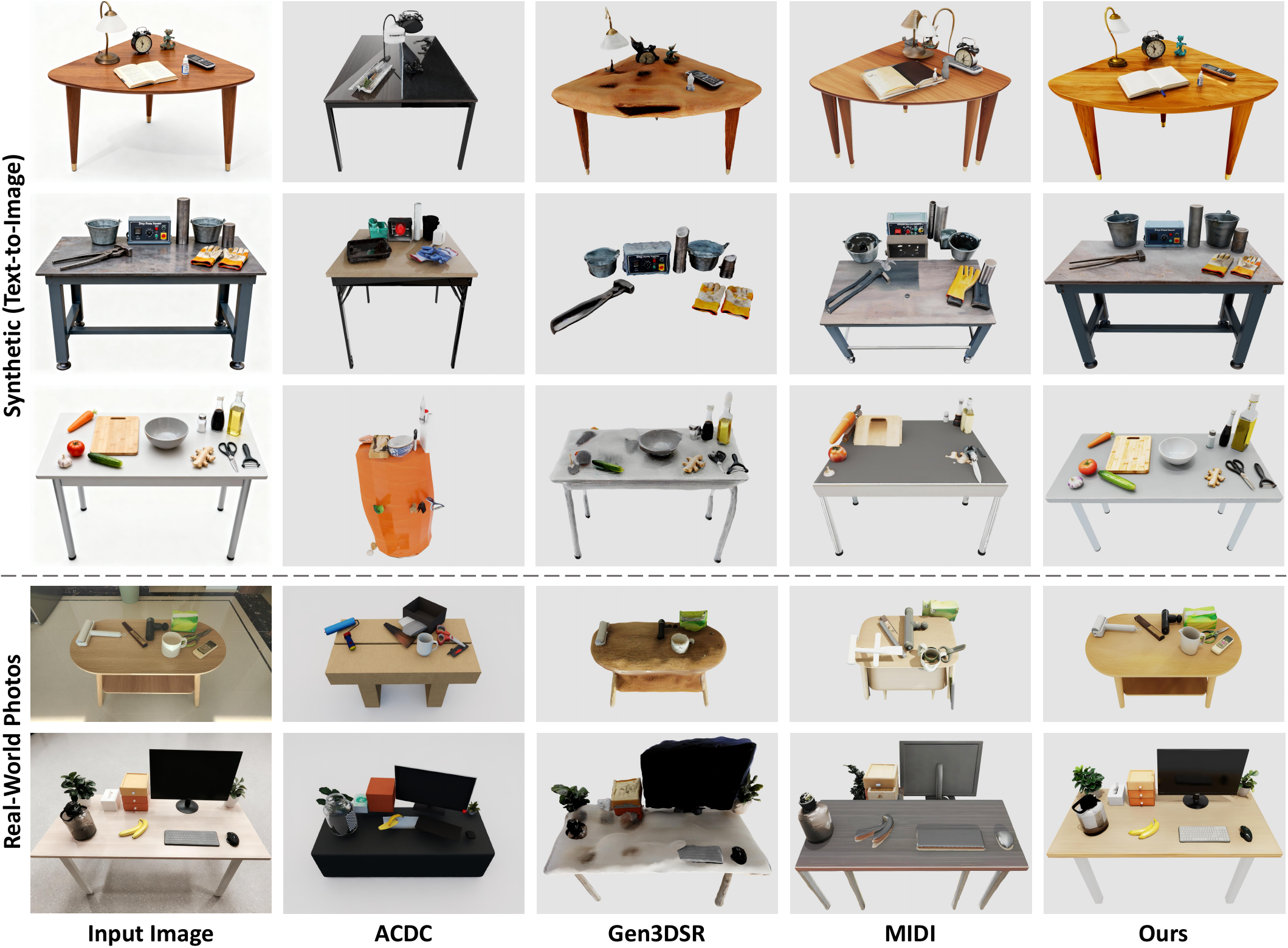}

  \caption{\textbf{Qualitative comparison on synthesized inputs (top) and real-world photos (bottom).} Our method, TabletopGen, consistently outperforms all baselines. TabletopGen demonstrates strong adaptability across diverse tabletop types, delivering more realistic appearances, finer instance models, more coherent object counts and layouts, and collision-free placement.}
  \label{fig:compare-image}
\end{figure}

\begin{figure}
  \centering
    \includegraphics[width=1.0\linewidth]{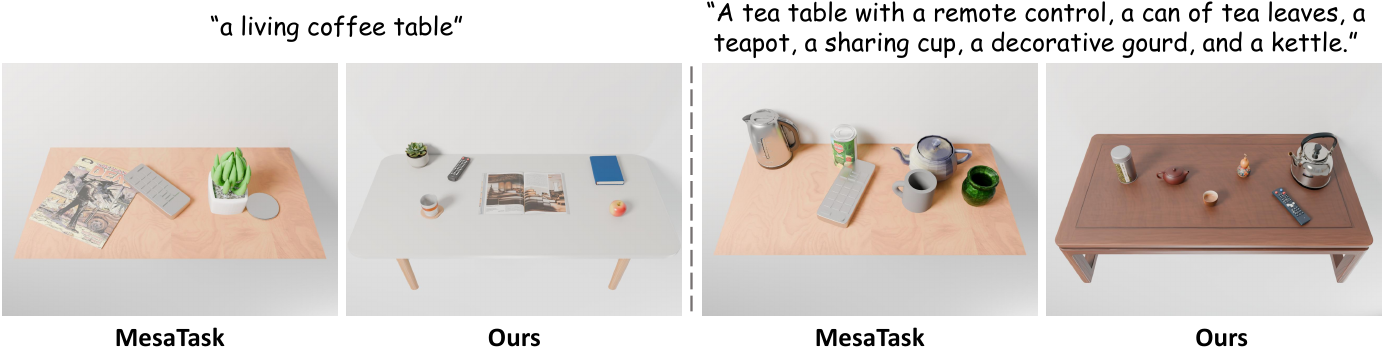}

   \caption{\textbf{Qualitative comparison under the same input texts.} Compared with the recent tabletop generation method MesaTask, TabletopGen generates more complete and realistic scenes, including stylistically consistent tables, more detailed instance models, and more semantically and physically reasonable layouts.}
   \label{fig:mesa}
\end{figure}

\subsubsection{Qualitative Evaluation.}
\cref{fig:compare-image} compares our method with \textbf{single-image} baselines on both synthesized images and real-world photos. Thanks to our precise per-instance generation and robust pose and scale alignment, TabletopGen produces consistent object counts and categories, finer geometry with unified style, semantically coherent layouts, and collision-free placement. In contrast, the retrieval-based method (\eg, ACDC) is limited by its fixed asset library, failing to match specific object styles and shapes. Generative reconstruction methods (\eg, Gen3DSR and MIDI) struggle with occlusions, causing incomplete instance generation, object interpenetration, and layout inaccuracies. This finding aligns with the high collision rates reported in \cref{tab:quantitative}. Overall, TabletopGen delivers more realistic, simulation-ready tabletops with diverse types.

We also demonstrate our \textbf{text-to-scene} capability by comparing with MesaTask in \cref{fig:mesa}. 
MesaTask retrieves assets and places them on a fixed plane, limiting diversity and fidelity and occasionally causing minor collisions(\eg, the remote control and the can in the bottom row).
In contrast, TabletopGen generates the entire scene, including the stylistically consistent table, delivering more realistic visual appearance, richer instance counts, more reasonable arrangements, and collision-free layouts.

\subsection{Ablations}

\begin{figure}[t]
  \centering
  \includegraphics[width=0.98\linewidth]{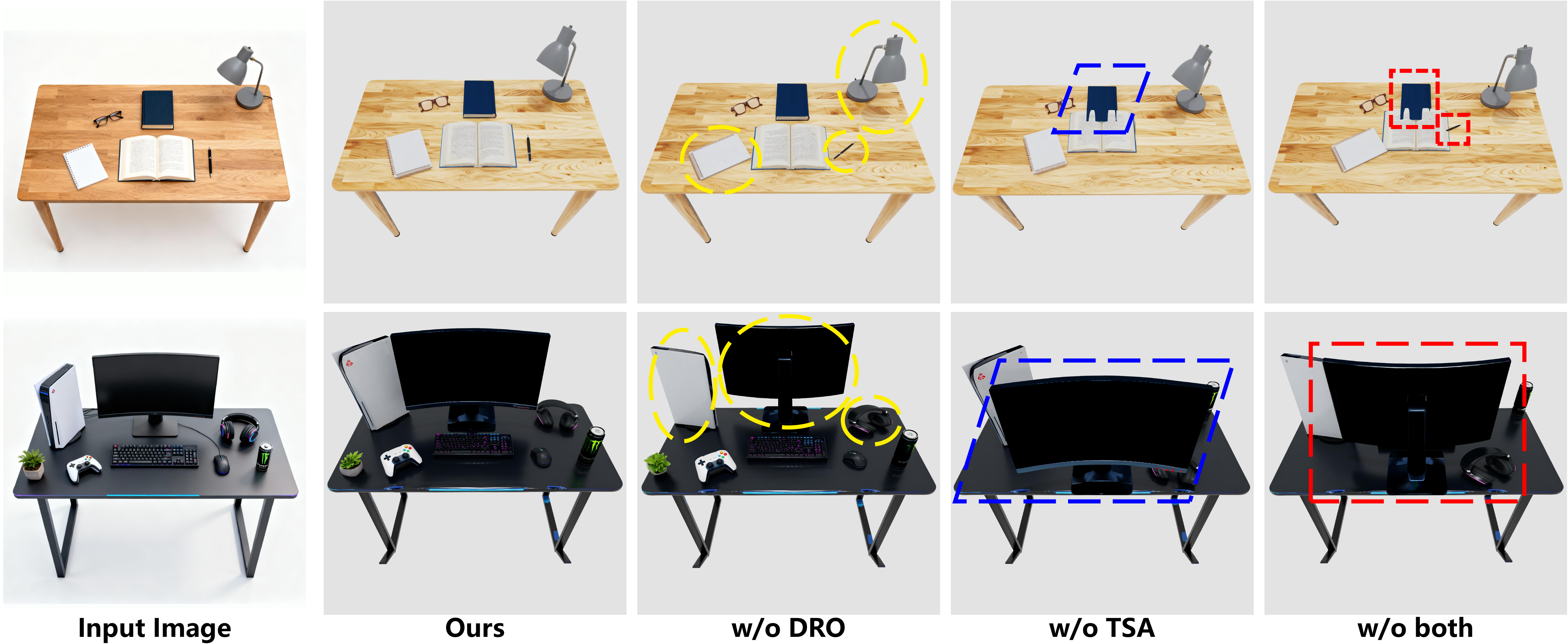}

  \caption{\textbf{Qualitative ablation study on pose and scale alignment components.} Compared to our full model (Ours), removing DRO yields incorrect rotations (\textcolor{yellow}{yellow circles}), removing TSA causes misplacements (\textcolor{blue}{blue parallelograms}), and removing both amplifies these errors, resulting in severe occlusions and collisions (\textcolor{red}{red rectangles}). }
  \label{fig:ablation}
\end{figure}

We ablate the two key stages, \textbf{DRO} and \textbf{TSA}, of the pose and scale alignment approach by replacing them with a naive MLLM-only baseline: when a stage is removed, we replace its specialized function by directly prompting ChatGPT to estimate the parameters from the reference image. As reported in \cref{tab:ablation}, removing either stage degrades visual \& perceptual scores and sharply increases collisions, while removing both leads to the largest failure. \cref{fig:ablation} visually confirms these findings. The "w/o DRO" baseline fails to recover correct rotations, while the "w/o TSA" baseline produces implausible placements and scale drift. The "w/o both" compounds these errors, causing the final scene to suffer from severe collisions (\eg, the books interpenetrating) and occlusions (\eg, the monitor moving forward and blocking other objects). In contrast, our full model's result (Ours) closely matches the input image and presents a collision-free layout. This demonstrates that both DRO and TSA are essential for achieving a coherent and physically plausible scene reconstruction.

\begin{table}
  \caption{\textbf{Quantitative ablation study on pose and scale alignment components.} Our full model (DRO+TSA) attains the best visual \& perceptual scores and the lowest collision rates; removing either component degrades performance, and removing both leads to large increases in collisions.}
  \label{tab:ablation}
  \setlength{\tabcolsep}{3pt}
  \centering
  \begin{tabular}{lccccc}
    \toprule \multirow{2}{*}{\textbf{Method}} & \multicolumn{3}{c}{\textbf{Visual \& Perceptual}} & \multicolumn{2}{c}{\textbf{Collision Rate (\%)}} \\
    \cmidrule(lr){2-4}\cmidrule(l){5-6}       & \textbf{LPIPS$\downarrow$}                        & \textbf{DINOv2$\uparrow$}                       & \textbf{CLIP$\uparrow$} & \textbf{Col\_O$\downarrow$} & \textbf{Col\_S$\downarrow$} \\
    \midrule Ours (full)                      & \textbf{0.4483}                                   & \textbf{0.8383}                                 & \textbf{0.9077}         & \textbf{0.42}               & \textbf{7.69}               \\
    w/o DRO                                   & 0.4523                                            & 0.8261                                          & 0.9012                  & 1.27                        & 16.67                       \\
    w/o TSA                                   & 0.4799                                            & 0.8041                                          & 0.8954                  & 5.50                        & 61.54                       \\
    w/o both                                  & 0.4811                                            & 0.7897                                          & 0.8922                  & 5.41                        & 62.82                       \\
    \bottomrule
  \end{tabular}
\end{table}

\subsection{Data Engine Demonstrations}

\subsubsection{Zero-shot Real-to-Sim-to-Real Policy Transfer.}

\begin{figure}[t]
  \centering
  \includegraphics[width=1.0\linewidth]{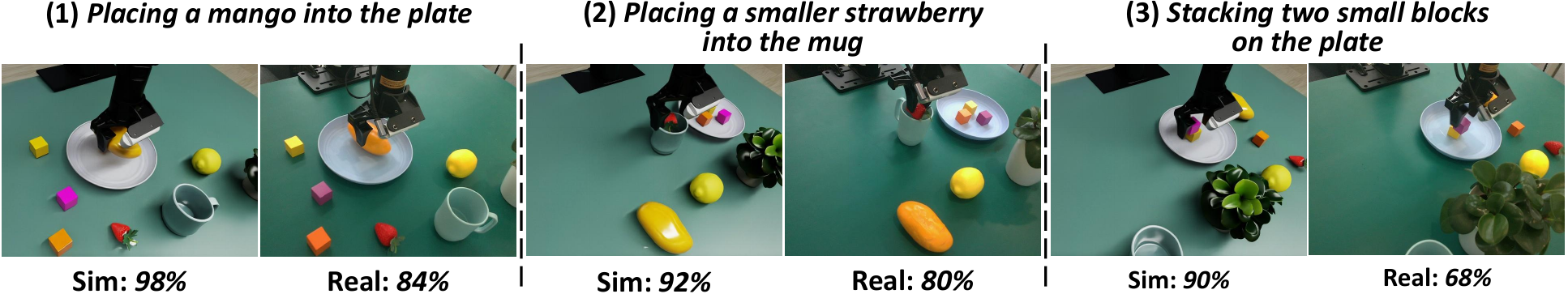}

  \caption{\textbf{Zero-shot real-to-sim-to-real policy learning.} In a simulated scene reconstructed from a real tabletop photo, we synthesize 2,000 trajectories for each of three tasks with different difficulty levels to fine-tune a $\pi_{0.5}$ policy, and directly deploy it on a physical AgileX PiPER arm without any real-world data fine-tuning. The high success rates in both simulation and the real world indicate that TabletopGen can generate high-quality and usable data for robot manipulation policy learning.}
  \label{fig:sim2real}
\end{figure}

To further validate that TabletopGen can support downstream policy learning, we train manipulation policies using only synthetic demonstrations and directly transfer them to the real world. Starting from a real tabletop photo containing fruits, containers, and distractor objects, we generate an instance-level simulated scene and attach collision geometry and physical properties. A motion planner is then used to compute collision-free trajectories for an AgileX PiPER arm to execute pick-and-place tasks in simulation. We consider three tasks with increasing difficulty: placing a mango onto a plate, placing a smaller strawberry into a mug, and stacking two small blocks on a plate. For each task, we apply domain randomization to the background lighting, arm pose, camera pose and focal length, and object poses, and synthesize 2,000 trajectories to fine-tune a $\pi_{0.5}$ policy~\cite{intelligence2025pi_}, without using any real-world data for fine-tuning. We then evaluate each task for 50 trials in both simulation and the original real-world scene. As shown in \cref{fig:sim2real}, the resulting policies achieve high success rates in both simulated and real-world executions. The larger sim-to-real gap in Task 3 is expected, since stacking is highly sensitive to friction, contact modeling, and release timing, where small errors may cause the blocks to slip or collapse. Overall, these results show that TabletopGen can provide realistic and usable synthetic data for sim-to-real robot manipulation policy learning.

\begin{figure}[t]
  \centering
  \includegraphics[width=1.0\linewidth]{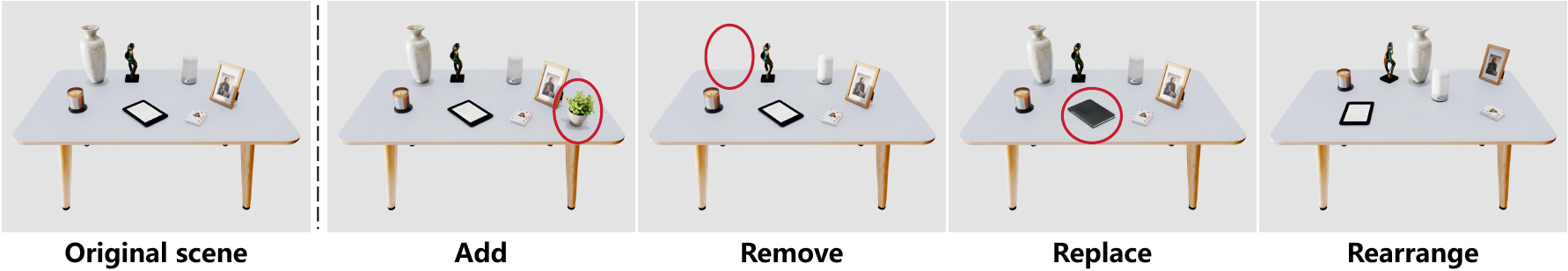}

  \caption{\textbf{In-scene Editing.} Starting from the original generated scene, we demonstrate four types of in-scene editing: Add, Remove, Replace, and Rearrange. These edits can quickly derive multiple scene variants, providing broader environment coverage for subsequent task generation and interactive data synthesis.}
  \label{fig:editing}
\end{figure}

\subsubsection{In-scene Editing for Scene Expansion.}
A key capability of a strong data engine is to efficiently expand data diversity. Since TabletopGen constructs scenes from independent 3D instances, the generated simulation environments are naturally editable and extensible. As shown in \cref{fig:editing}, we derive multiple scene variants by adding, removing, replacing, or rearranging objects, without re-running the full pipeline. This editing preserves the overall scene semantics while producing diverse simulation environments, thereby broadening the coverage of downstream task generation and interaction data synthesis at a low additional cost.

%% file: sec/5_conclusion.tex
\section{Conclusion}
\label{sec:conclusion}

In this paper, we present \textbf{TabletopGen}, a training-free, fully automatic tabletop scene generation and interactive simulation engine. It converts text or a single image into high-fidelity, instance-level interactable, and physically plausible 3D tabletop scenes, which can be utilized for multimodal data synthesis in robotic manipulation tasks. We decouple 2D-to-3D layout recovery into two stages: a Differentiable Rotation Optimizer (DRO) to accurately recover per-instance orientations, and a Top-view Spatial Alignment (TSA) module that injects tabletop-specific physical priors to infer globally consistent translations and metric scales, yielding collision-free and executable layouts. Extensive quantitative evaluations and a large-scale user study show that TabletopGen achieves state-of-the-art performance in visual quality, layout accuracy, and physical validity. Moreover, we demonstrate zero-shot real-to-sim-to-real policy transfer, showing that policies fine-tuned only on synthetic trajectories generated by TabletopGen can be directly deployed for real-world robotic manipulation. Finally, thanks to efficient in-scene editing and text-driven scene expansion, TabletopGen provides a scalable approach to deriving diverse simulation environments at low cost, offering a practical route to alleviate the data bottleneck in embodied AI.

\subsection{Limitations and Future Work.} TabletopGen currently focuses on typical tabletop scenes, and future work can explore extending it to more complex support structures, such as multi-level shelves and cabinets. Moreover, given the data-engine capability demonstrated by TabletopGen, our future focus will be to use this framework to synthesize large-scale, highly diverse simulated interaction data for policy training, thereby enhancing robots' generalization ability in complex real-world manipulation tasks.

%% file: sec/supp.tex
\clearpage
\begin{center}
  {\Large\bfseries Supplementary Material for TabletopGen\par}
\end{center}
\vspace{1.5em}

This is the Supplementary Material for \textbf{TabletopGen}. We provide additional qualitative results, video demonstration descriptions, and comprehensive experimental configurations that complement the main text. Specifically, \cref{add-qual} presents further qualitative results, including a gallery of diverse generated tabletop scenes, detailed stage-wise intermediate outputs of our pipeline, and extended comparisons with recent scene generation methods. In \cref{video}, we introduce our supplementary videos, which showcase the interactive nature of the generated simulation environments and validate our zero-shot real-to-sim-to-real transfer capabilities. 
Finally, \cref{detail-ex} provides detailed experimental and implementation setups, covering the construction of our test set, the standardized camera sweep protocol used for fair baseline comparisons, comprehensive user study execution details, and the prompt templates used across all stages of TabletopGen.

\begin{figure}[h]
  \centering
   \includegraphics[width=1.0\linewidth]{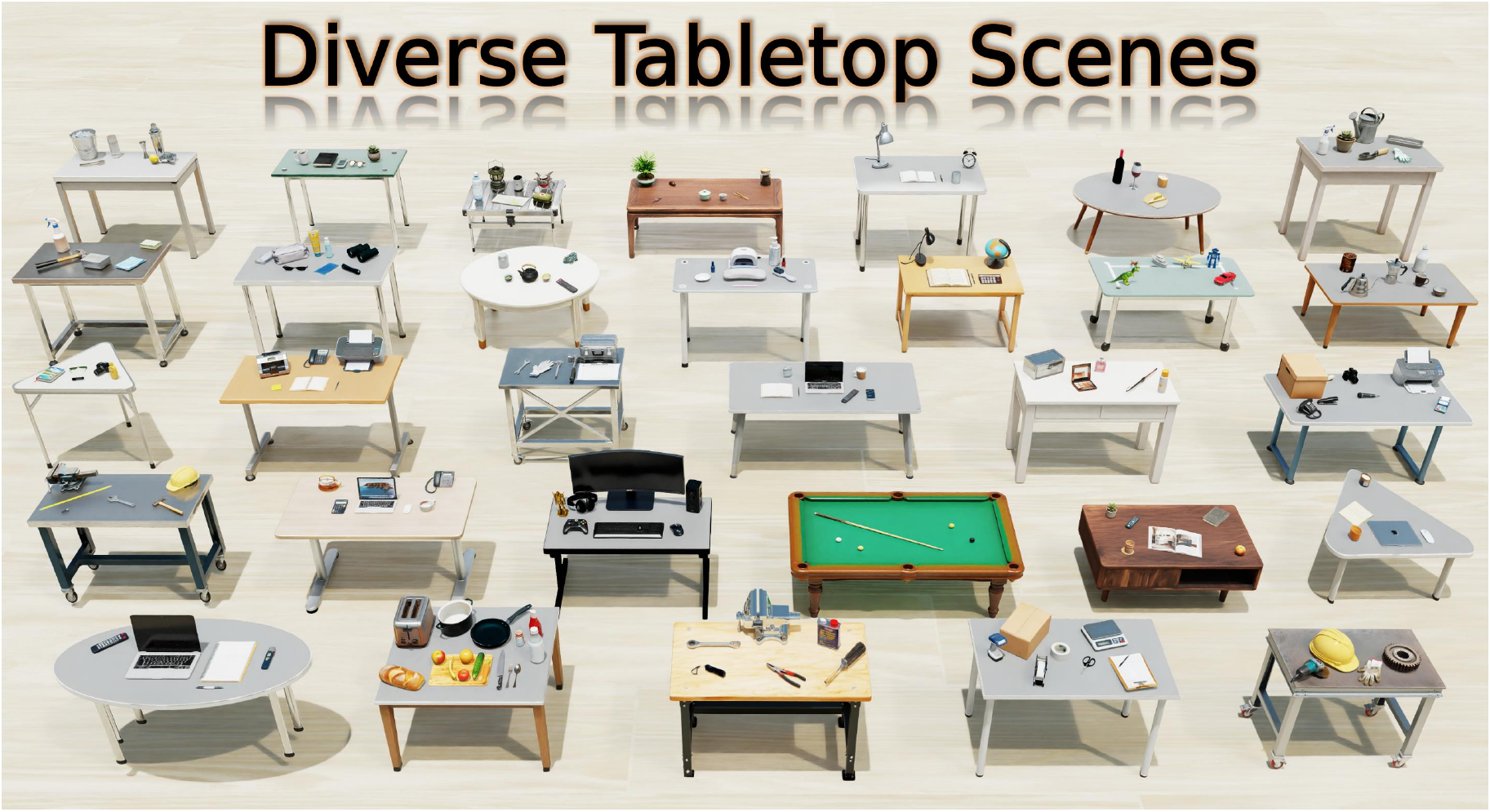}
   \caption{\textbf{A comprehensive gallery of diverse tabletop scenes generated by TabletopGen.} This panoramic overview showcases a wide array of tabletops (\eg, office desks, coffee tables, dining tables, workbenches, and various other types) populated with domain-specific objects. Every scene presented here is equipped with collision geometries and physical properties, ready for immediate deployment in interactive physics simulations and large-scale data synthesis.}
   \label{fig:gallery}
\end{figure}

\section{Additional Qualitative Results}
\label{add-qual}

\subsection{Scene Gallery}

To comprehensively demonstrate the diversity, scalability, and robust generation capabilities of TabletopGen, we first present a large-scale panoramic gallery of our generated scenes (\cref{fig:gallery}). These results emphasize our framework's ability to handle various tabletop types and complex instance layouts.

\subsection{Stage-wise Intermediate Outputs}

TabletopGen is a multi-stage 3D scene generation framework that decomposes the complex task of converting text or a single image into a 3D scene into several more manageable sub-stages, including instance extraction, canonical 3D model generation, pose and scale alignment, and final 3D scene assembly. To enhance transparency and reproducibility, we present stage-wise key intermediate results from the inputs to the final outputs.

\subsubsection{Reference Image Synthesis.} 
\begin{figure}[h]
  \centering
   \includegraphics[width=1.0\linewidth]{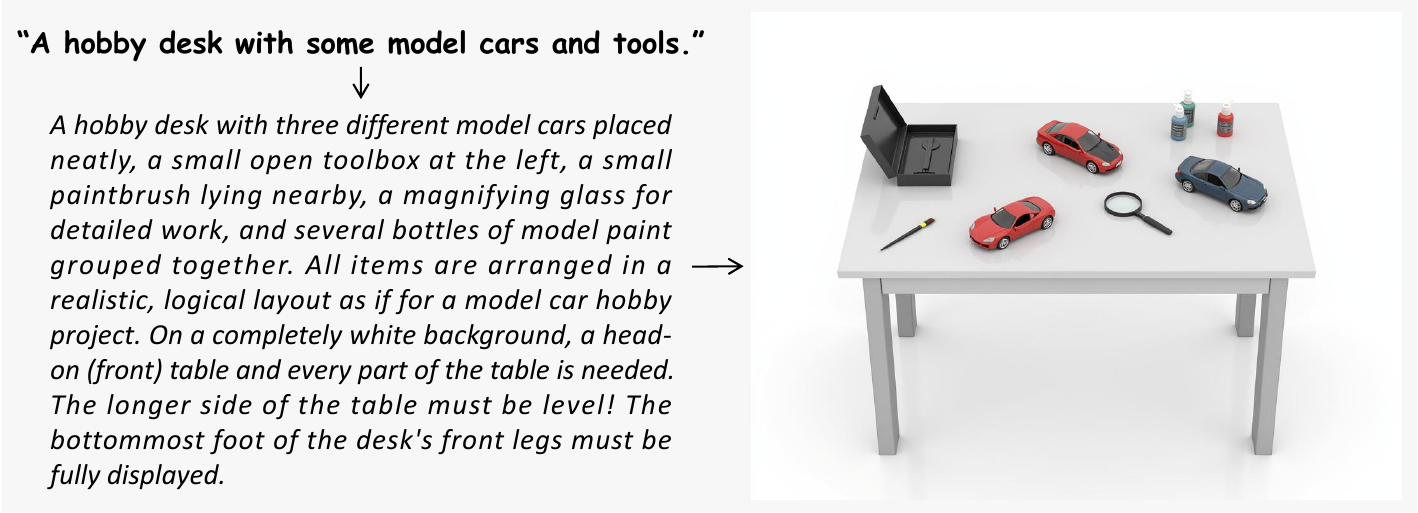}
   \caption{\textbf{Text-input preprocessing.} For a short text, we first use an LLM to expand it into a detailed, layout-aware scene description (left). The expanded description is then fed into a text-to-image model to synthesize the reference image (right), which conditions all subsequent stages.}
   \label{fig:t-t-i}
\end{figure}

We use a text input as an illustrative example. As shown in \cref{fig:t-t-i}, for a short text input by the user (\eg, "A hobby desk with some model cars and tools."), we first use ChatGPT to expand it into a detailed scene description. This description is then passed into the image generation model Seedream to synthesize a visually realistic and plausibly arranged reference image $I_{ref}$, on which all subsequent generation stages are processed.

\begin{figure}[h]
  \centering
   \includegraphics[width=1.0\linewidth]{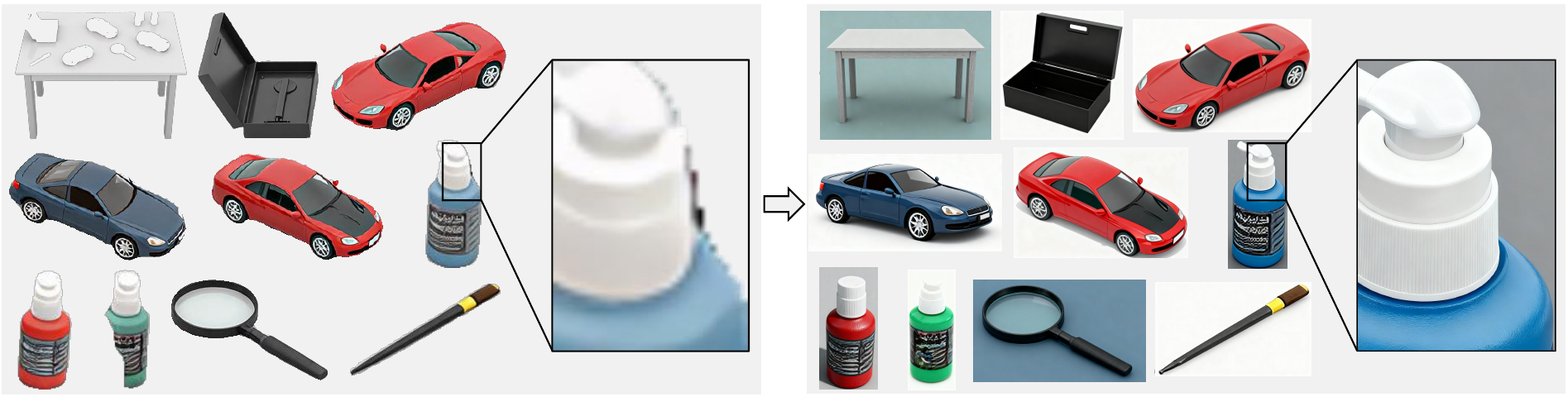}
   \caption{\textbf{Completion of segmentation masks.} We apply generative completion to refine the raw segmentation outputs: instance masks with holes and blurry boundaries are redrawn into complete, high-resolution, clean-edged instance images.}
   \label{fig:redraw}
\end{figure}

\subsubsection{Instance Completion.} Beginning with $I_{ref}$, ChatGPT first analyzes the image to identify all object categories $L_i$ (\eg, "table", "toy car", "magnifying glass"). These categories then guide GroundedSAM-v2 to generate instance segmentation masks $M_i$. However, as shown in \cref{fig:redraw}, the masks often contain holes and blurry boundaries. To address this, we use Seedream for per-instance generative completion, producing high-resolution, clean instance images.

\subsubsection{Canonical 3D Model Generation.} 
\begin{figure}[h]
  \centering
   \includegraphics[width=0.8\linewidth]{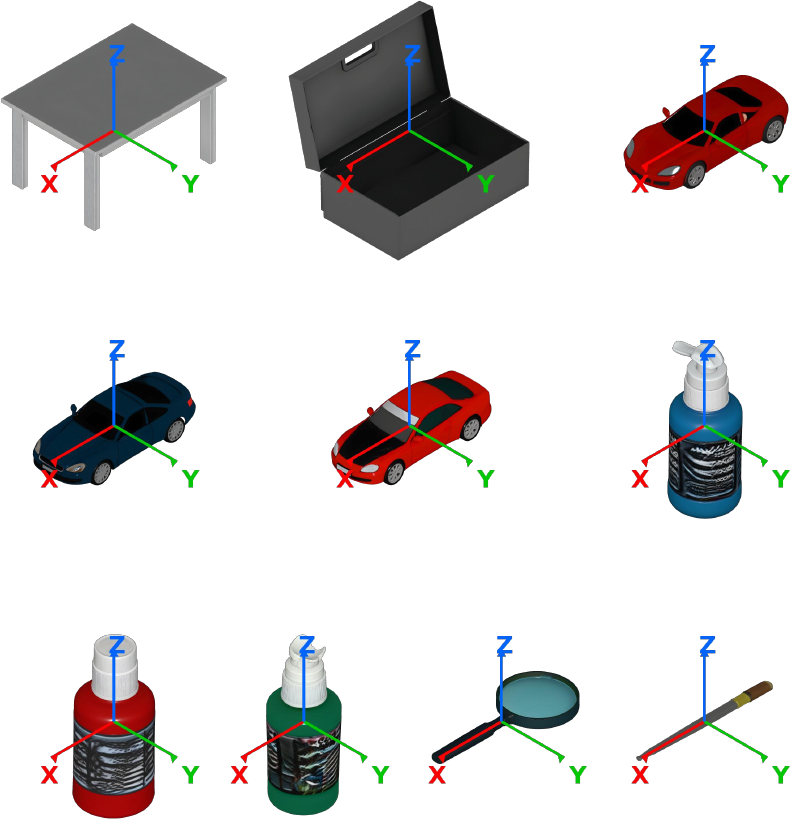}
   \caption{\textbf{Canonicalized instance 3D models.} After image-to-3D and coordinate alignment, each object's local vertical axis (+Z) is aligned with the tabletop world up, enabling direct placement.}
   \label{fig:canonical}
\end{figure}

From the completed instance images, we reconstruct high-quality 3D models using an image-to-3D model, Hunyuan3D-3.0. Subsequently, each initial model undergoes canonical coordinate alignment. We utilize ChatGPT to analyze whether its z-axis represents the correct upright orientation. If it does not, a corrective rotation is applied to align the model's local z-axis with the world coordinate system. \cref{fig:canonical} shows the outputs of this stage: all instance renderings in the canonical coordinates.

\subsubsection{Differentiable Rotation Optimizer.} After obtaining all 3D instances, we proceed to the Pose and Scale Alignment stage. This stage begins with the Differentiable Rotation Optimizer (DRO), which estimates the precise rotation $r_i$ for each instance in the scene. As described in Sec. \ref{sec:pose}, ChatGPT first predicts an initial camera perspective (azimuth, elevation) from the reference image $I_{ref}$. For our ``hobby desk'' example, this is (azimuth = 0$^\circ$, elevation = 60$^\circ$). Given this perspective, we differentiably render each instance and optimize its rotation $r_i$ by minimizing the tri-modal loss $\mathcal{L}_{rot}$. 

For supervision, we use the original instance crop with its segmentation mask, together with its silhouette and edge maps as reference images; all references are tightly cropped and resized to 256$\times$256. The rendered projections are cropped with the same tight bounds and resized to the same resolution to ensure loss consistency.

\begin{figure}[tb]
  \centering
  \begin{subfigure}[c]{0.6\linewidth}
    \centering
    \includegraphics[width=\linewidth]{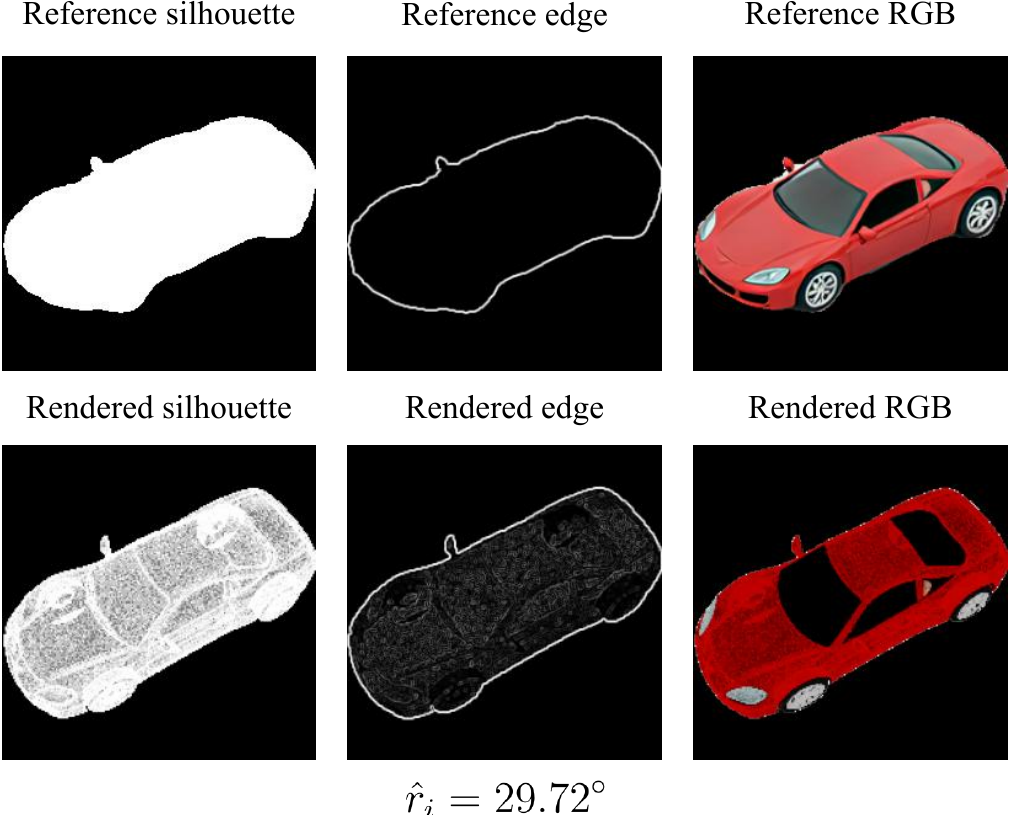}
    \caption{Results of the DRO for a ``toy car'' instance}
    \label{fig:rot_toycar2}
  \end{subfigure}
  \hfill
  \begin{subfigure}[c]{0.38\linewidth}
    \centering
    \includegraphics[width=\linewidth]{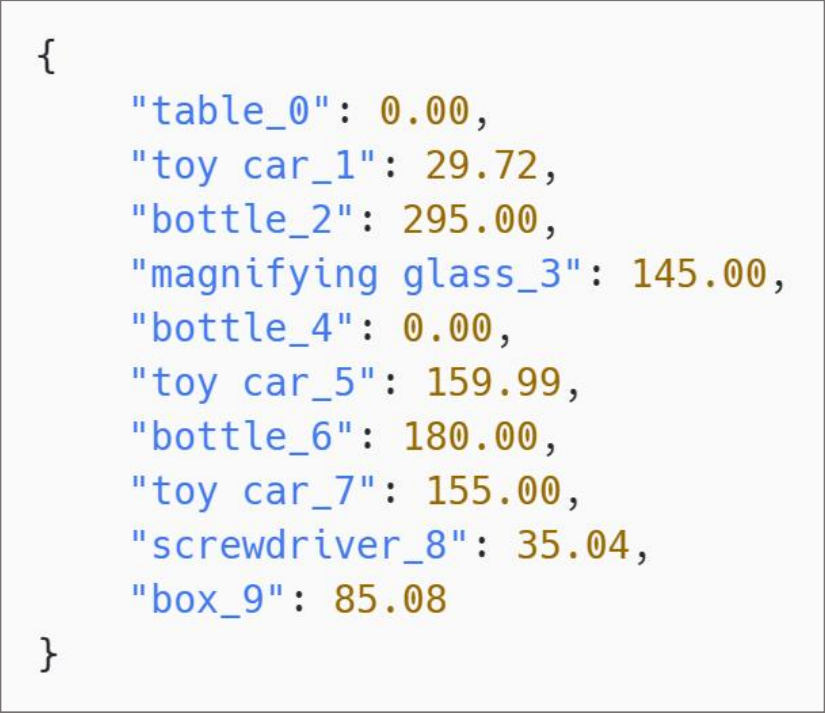}
    \caption{DRO outputs (JSON)}
    \label{fig:json_r}
  \end{subfigure}
  \caption{\textbf{Differentiable Rotation Optimizer (DRO) Results.} (a) The top row displays the reference silhouette, edge, and RGB maps. The bottom row shows the corresponding outputs from the differentiable renderer at the final estimated rotation ($\hat{r}_i = 29.72^\circ$). The high degree of consistency across all three modalities validates the effectiveness of our tri-modal loss in accurately recovering the instance's rotation. (b) Estimated per-instance rotation $r_i$ (Euler \emph{yaw}) about the scene $z$-axis, reported in degrees ($^\circ$).}
  \label{fig:dro_results}
\end{figure}

\cref{fig:rot_toycar2} shows the final match for instance \textit{toy car\_1}, where the rendered silhouette, edges, and appearance maps closely agree with the corresponding reference images. \cref{fig:json_r} shows the final JSON output of the rotation estimation.

\begin{figure}[tb]
  \centering
  \begin{subfigure}[c]{0.55\linewidth}
    \centering
    \includegraphics[width=\linewidth]{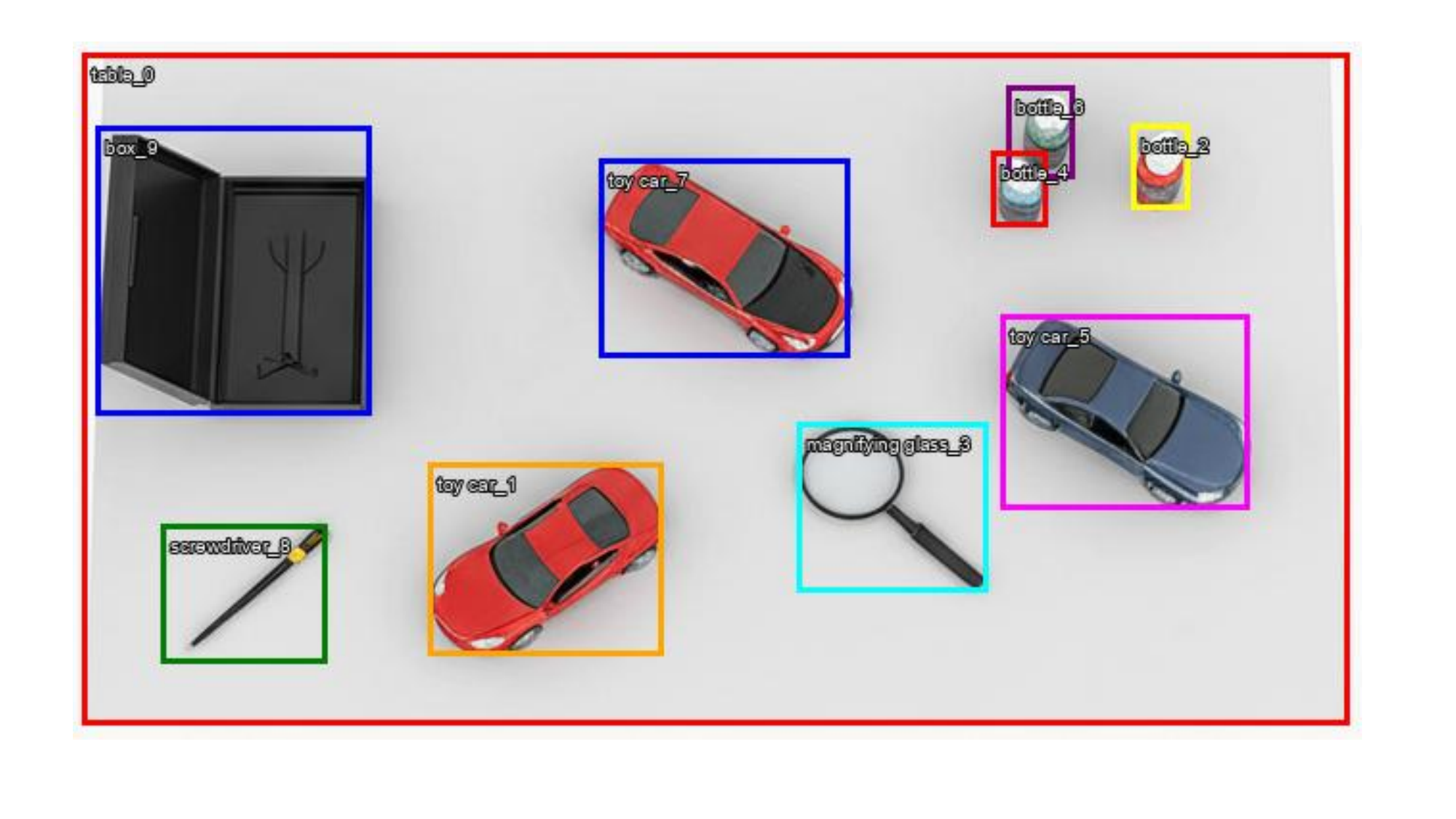}
    \caption{Top-view synthesis and bounding box detection}
    \label{fig:topview}
  \end{subfigure}
  \hfill
  \begin{subfigure}[c]{0.43\linewidth}
    \centering
    \includegraphics[width=\linewidth]{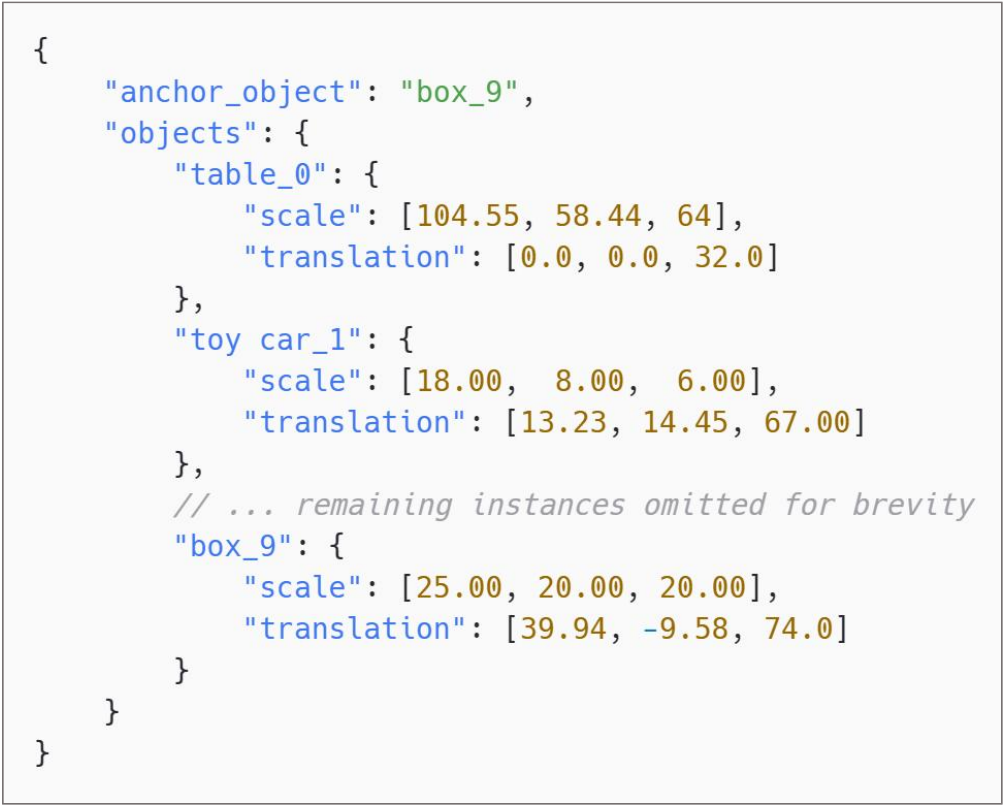}
    \caption{TSA outputs (JSON)}
    \label{fig:json_t}
  \end{subfigure}
  \caption{\textbf{Top-View Spatial Alignment (TSA) Results.} (a) A top-view of the tabletop is synthesized with a multimodal image generation model, and 2D bounding boxes for all instances are detected on it. These boxes are used to compute the RMA-Score for anchor selection and to derive translation and scale. (b) Estimated per-instance scale and translation, reported in centimeters (cm). Scales are listed as $[x,y,z]$ and translations as $[x,y,z]$ in the tabletop frame (origin at the table center; $+x$ left, $+y$ forward, $+z$ up). The anchor object selected by RMA-Score is shown at the top.}
  \label{fig:tsa_results}
\end{figure}

\subsubsection{Top-View Spatial Alignment.} The Top-View Spatial Alignment (TSA) estimates per-instance translation ($t_i$) and scale ($s_i$). We first synthesize a top-view image of the scene with Seedream and detect 2D bounding boxes for all instances in that view, as shown in \cref{fig:topview}. In parallel, ChatGPT provides commonsense physical sizes for each object. Subsequently, we identify the scene's anchor object (excluding the table) via the RMA-Score; in this example, it is \textit{box\_9}. Using this anchor, we rescale objects whose commonsense size deviates significantly from the bounding box size. We found experimentally that typically only the table requires scaling, as the MLLM tends to overestimate its size. 

After scaling, we define a coordinate system with the table's center as the origin, the horizontal-left direction as the positive x-axis, and the forward direction as the positive y-axis. The xy-translation for each instance is then calculated using the center point of its bounding box. To handle stacking situations (\eg, a pen on a book), we also use ChatGPT to analyze the stacking order of instances in the scene to determine the z-translation. \cref{fig:json_t} shows the final JSON output of the translation and scale estimation.

\subsubsection{Data Synthesis in Interactive Simulation.} Once all instances are aligned and assembled with collision geometries and physical properties in the simulator, TabletopGen functions as a scalable data engine. As the robot executes planned trajectories within these generated environments, we can synthesize highly diverse multimodal data. 

\begin{figure}[tb]
  \centering
  \includegraphics[width=1.0\linewidth]{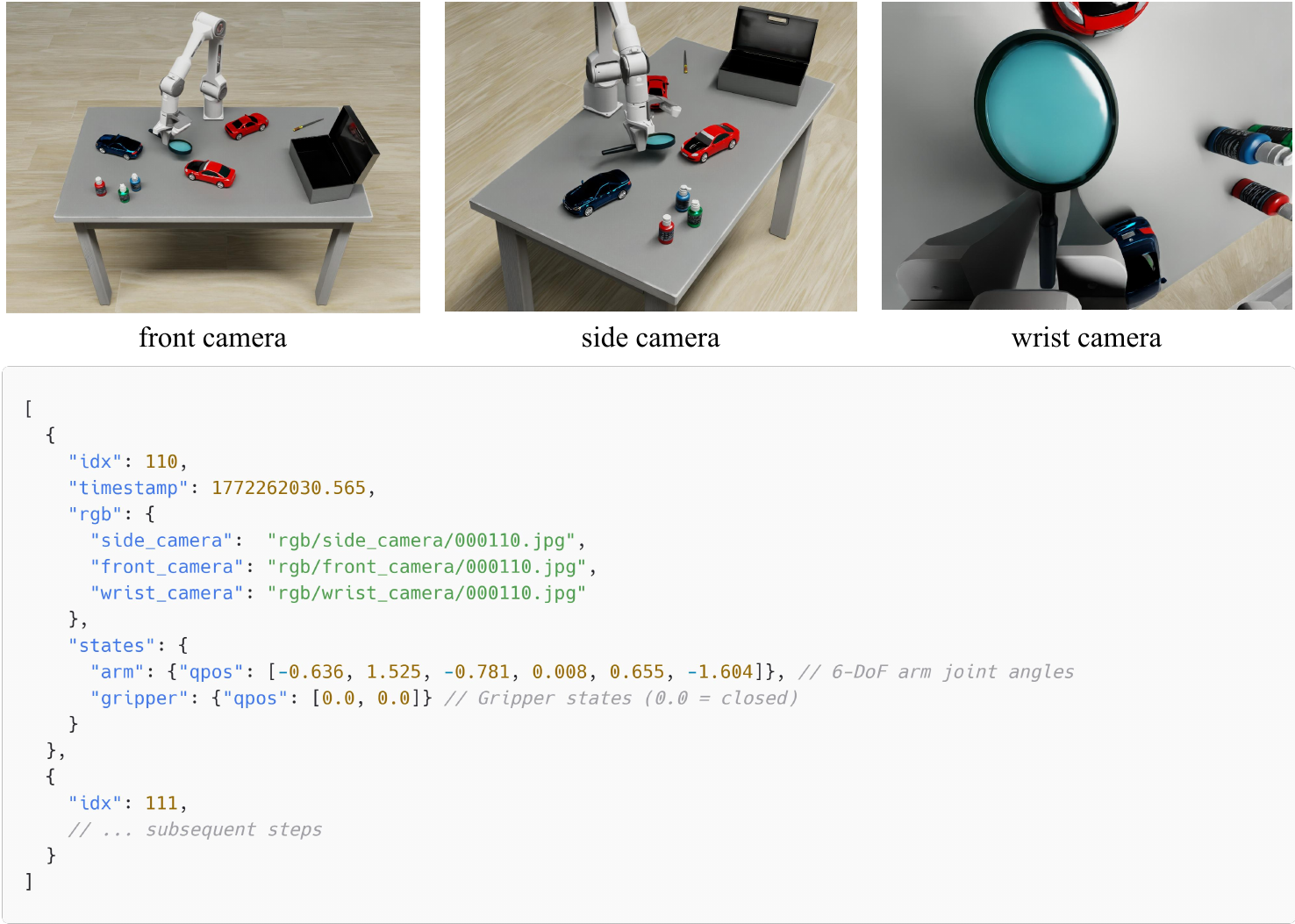}
  \caption{\textbf{Data synthesis output in interactive simulation.} Synchronized multi-view RGB observations, captured from the side, front, and wrist cameras during the robotic manipulation task, are displayed in the upper section. A snippet of the corresponding JSON trajectory data, which records temporal information and the robot's joint states, is shown below. Our framework is highly extensible and can be effortlessly configured to collect various other modalities as needed, such as depth maps, segmentation masks, and point clouds.}
  \label{fig:sim_data}
\end{figure}

To demonstrate the capability of our data engine, \cref{fig:sim_data} illustrates a typical collected data frame. The basic configuration captures synchronized multi-view RGB observations (\eg, side, front, and wrist cameras) alongside the robot's joint states, which can be directly utilized to compute execution actions. Thanks to the flexibility of our simulation pipeline, researchers can easily configure the simulator to output a wider variety of rich modalities according to downstream training requirements, such as depth maps, instance segmentation masks, and precise point clouds.

\subsection{Comparison with Recent Methods}

We further compare TabletopGen with two recent scene generation methods, SceneGen \cite{meng2025scenegen} and FlashSculptor \cite{hu2025flashsculptormodular3d}, and include MIDI as the strongest baseline from the main paper for reference. In addition to the visual metrics and mesh-level collision rates used in the main paper, we report two physical reliability metrics. Support Stability (SS) measures the percentage of objects whose projected center of mass lies within a valid support region, reflecting quasi-static support validity. Dynamic Settling Stability (DSS) imports the generated scene into simulation, enables gravity and collisions, lets the scene settle for 5 seconds, and counts objects whose translation changes by less than 3\,cm and rotation changes by less than $30^\circ$.

As shown in \cref{tab:recent_methods,fig:recent_methods}, TabletopGen achieves the best visual scores and substantially higher physical reliability than the compared methods. In particular, SceneGen and FlashSculptor can produce visually recognizable scenes, but their layouts often contain unsupported objects, objects outside the tabletop boundary, or severe interpenetration. By contrast, TabletopGen maintains near-zero collision rates while achieving 100.00\% SS and 92.13\% DSS, indicating that its generated tabletop scenes are not only visually consistent but also directly usable for physics-based simulation.

\begin{center}
  \captionof{table}{\textbf{Comparison with recent scene generation methods.} TabletopGen achieves the best visual scores and substantially better physical stability.}
  \label{tab:recent_methods}
  \setlength{\tabcolsep}{4pt}
  \small
  \begin{tabular}{@{}lccccccc@{}}
    \toprule
    \multirow{2}{*}{\textbf{Method}} & \multicolumn{3}{c}{\textbf{Visual}} & \multicolumn{4}{c}{\textbf{Physical Stability (\%)}} \\
    \cmidrule(lr){2-4} \cmidrule(l){5-8}
    & LPIPS$\downarrow$ & DINOv2$\uparrow$ & CLIP$\uparrow$ & Col\_O$\downarrow$ & Col\_S$\downarrow$ & SS$\uparrow$ & DSS$\uparrow$ \\
    \midrule
    MIDI           & 0.4559 & 0.7070 & 0.8867 & 17.39 & 98.72 & 45.11  & 3.96 \\
    SceneGen       & 0.5326 & 0.6253 & 0.8397 & 21.69 & 98.72 & 4.43   & 0.67 \\
    FlashSculptor  & 0.5211 & 0.4636 & 0.8003 & 9.11  & 78.21 & 10.28  & 6.42 \\
    Ours           & \textbf{0.4483} & \textbf{0.8383} & \textbf{0.9077} & \textbf{0.42} & \textbf{7.69} & \textbf{100.00} & \textbf{92.13} \\
    \bottomrule
  \end{tabular}
\end{center}

\begin{center}
  \includegraphics[width=1.0\linewidth]{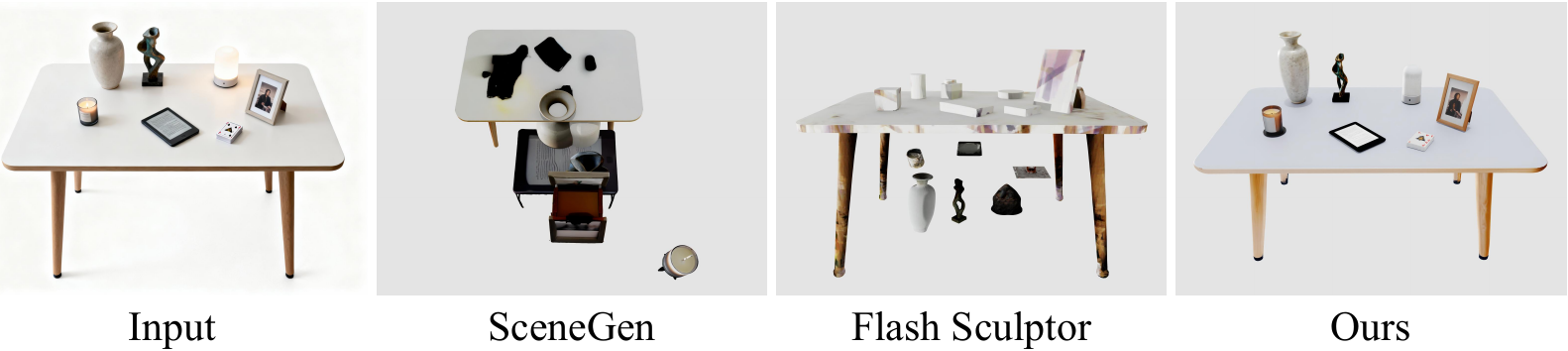}
  \captionof{figure}{\textbf{Qualitative comparison with recent methods.} Compared with SceneGen and FlashSculptor, TabletopGen better preserves object scale, tabletop support, and coherent spatial arrangement, producing a reconstruction that is visually aligned with the input and more physically plausible.}
  \label{fig:recent_methods}
\end{center}

\section{Video Demonstrations} 
\label{video}

\subsection{Interactive Scene Showcase}

To provide a more detailed and intuitive demonstration of the visual fidelity and interactive capability of our generated scenes, the supplementary video, "\textit{SceneShowcase.mp4}", presents a $360^\circ$ orbital demonstration of multiple scenes. The video showcases the scenes' structural integrity via white model visualization (geometry) and their final full texture rendering. Furthermore, we actively demonstrate the instance-level interactive property by executing movement on individual objects within each scene. This confirms the physical viability of every generated scene, highlighting the application potential of TabletopGen in physics-enabled simulation environments.

\subsection{Robotic Manipulation Demonstrations}

We utilize the NVIDIA Isaac Sim 4.5.0 environment to place the generated tabletops into an indoor scene, and introduce a Franka Emika Panda 7-DoF arm with a parallel gripper into the scene. This widely used, standard industrial arm is tasked with executing a series of complex pick-and-place maneuvers to demonstrate the physical viability and suitability of our scenes for embodied manipulation tasks.

We selected two common tabletop environments, a kitchen table and a coffee table, for demonstration:
\begin{itemize}
    \item Kitchen Task: The arm was tasked with executing a multi-step food preparation sequence: placing the cucumber into the bowl, and subsequently moving the carrot and salt shaker onto the cutting board.
    
    \item Coffee Table Task: The task required the arm to swap the positions of the cup and the apple.
\end{itemize}

The accompanying supplementary video, "\textit{ManipulationTaskDemo.mp4}", illustrates the successful execution of both tasks without spurious penetrations or unstable contacts. This demonstrates that TabletopGen produces interaction-ready tabletop scenes whose instance models and physical properties are directly usable for embodied AI environments and robot manipulation policy learning.

\subsection{Zero-Shot Real-to-Sim-to-Real Trajectory Transfer} 
\begin{figure}[tb]
  \centering
  \includegraphics[width=1.0\linewidth]{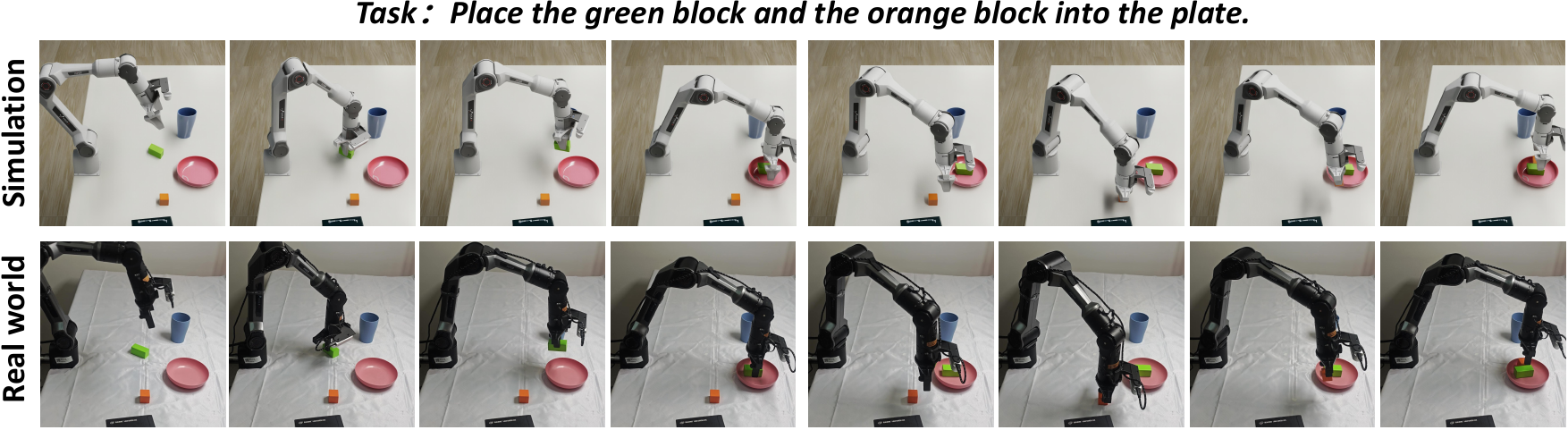}
  \caption{\textbf{Zero-shot real-to-sim-to-real trajectory transfer.} We plan a pick-and-place trajectory in the simulated scene reconstructed from a real tabletop photo and replay it on a physical AgileX PiPER arm without fine-tuning, successfully completing the same task. This complementary experiment validates the spatial and physical executability of trajectories synthesized in TabletopGen scenes.}
  \label{fig:si_sim_real_transfer}
\end{figure}

To further validate the physical executability of TabletopGen scenes, we provide a full continuous demonstration of zero-shot real-to-sim-to-real trajectory transfer in the supplementary video ``\textit{Sim2Real\_Transfer.mp4}'' and \cref{fig:si_sim_real_transfer}.

Specifically, we directly apply the sequence of trajectory waypoints planned in our simulated scene to the real physical robot without any fine-tuning. The synchronized side-by-side video confirms that the real robot successfully completes the pick-and-place task, achieving highly consistent behaviors with the simulation. 

Please note that since the primary objective of this demonstration is to validate the spatial and physical viability of the synthesized trajectories, the real-world camera viewpoint was not strictly aligned with the simulation camera.

\section{Details of Experiments}
\label{detail-ex}

\subsection{Test Set Construction}
Our test set contains 78 tabletop scenes, including 74 synthesized images and 4 real-world photographs. To ensure broad coverage, we design the set along two axes. First, for functional diversity, we follow the MesaTask taxonomy and common tabletop scenarios suggested by the language model, covering office/living-room tables, dining/kitchen tables, workbench/utility tables, and crafting/hobby tables, with 31, 21, 14, and 12 samples respectively. The number of samples in each category is chosen to roughly reflect its prevalence in real-world tabletop scenarios, so that common household scenes are well represented while less frequent but manipulation-relevant settings are still included. Second, for geometric diversity, in addition to rectangular tables (66 samples), we include round/oval tables (8 samples) and triangular/irregular tables (4 samples), stressing each method under different tabletop geometries rather than only rectangular planes.

\begin{figure}[!t]
  \centering
  \begin{subfigure}{0.95\linewidth}
    \centering
    \includegraphics[width=\linewidth]{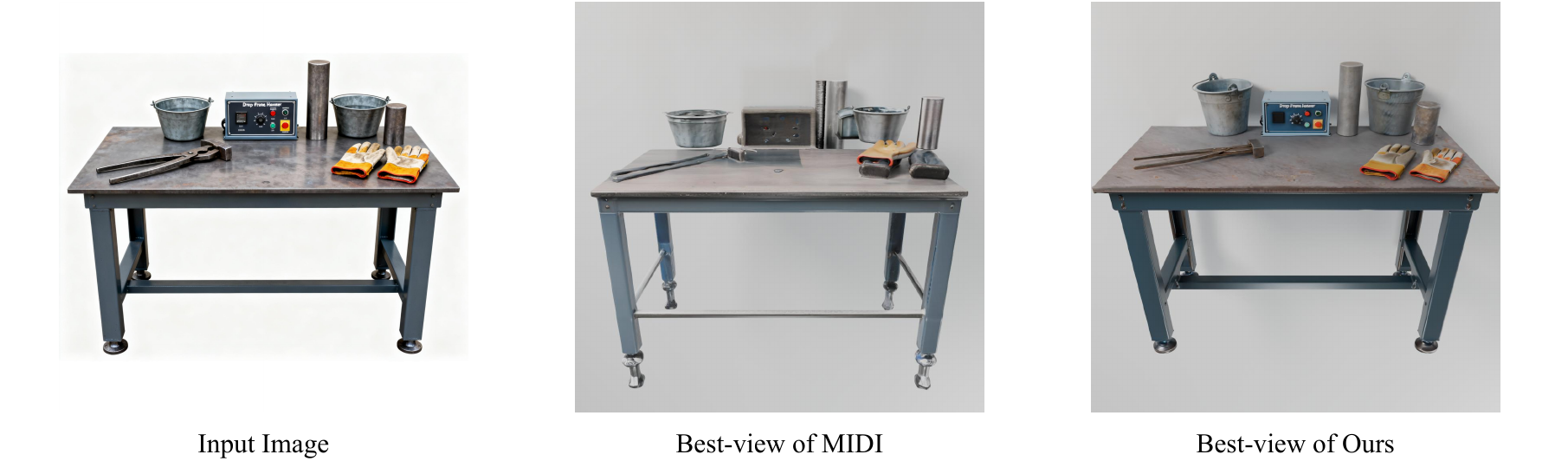}
    \caption{\textbf{Best views selected by LPIPS$\downarrow$.} 
    For each method, we render 160 views and pick the one with the lowest LPIPS$\downarrow$ to the input image.}
    \label{fig:lpips-best}
  \end{subfigure}

  \begin{subfigure}{0.95\linewidth}
    \centering
    \includegraphics[width=\linewidth]{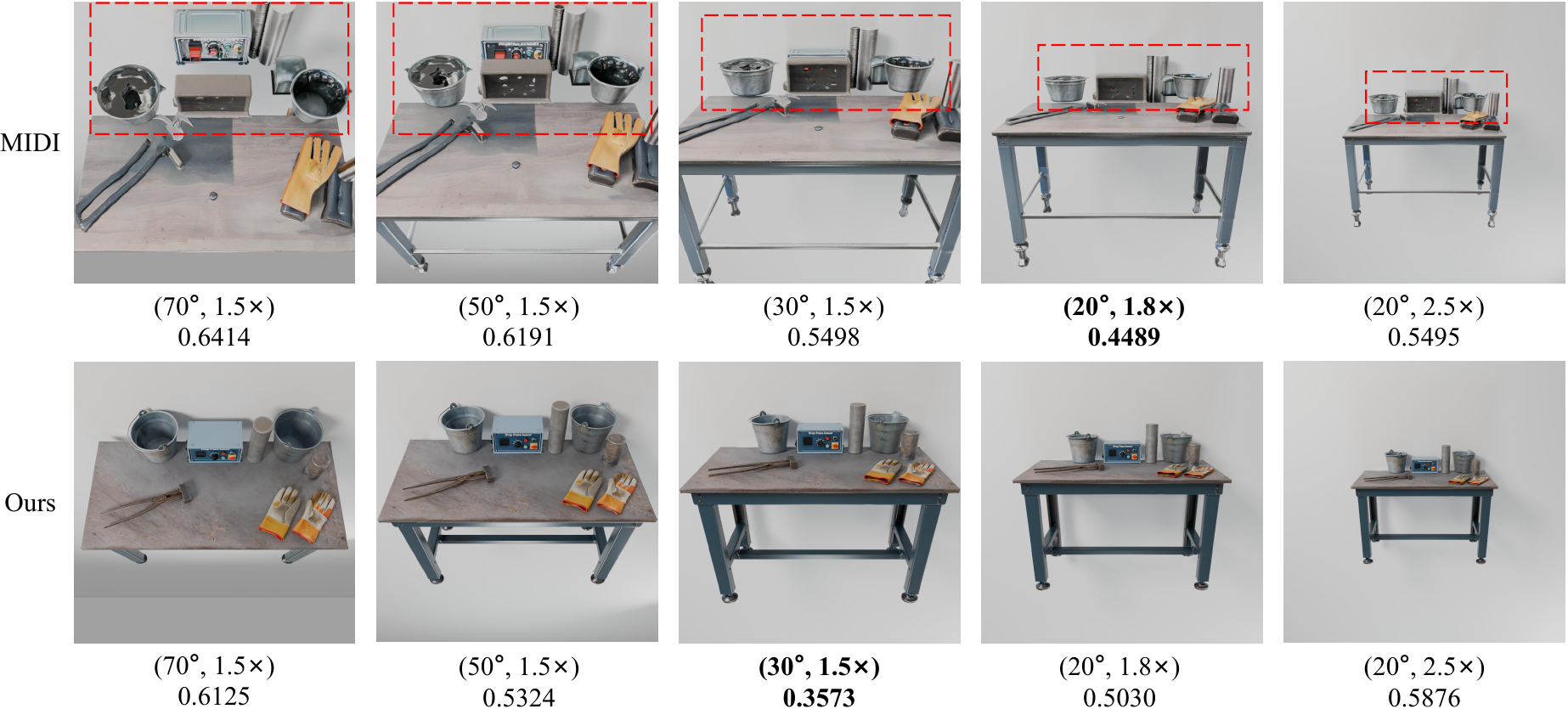}
    \caption{\textbf{LPIPS across different views.} Each image is annotated by (elevation$^\circ$, distance-multiplier) and its LPIPS$\downarrow$. 
    \textbf{Bold} denotes the best LPIPS$\downarrow$ / best view for that method. 
    \textcolor{red}{Red dashed rectangles} highlight the visual bias where visually plausible shots can hide physically invalid layouts (\eg, floating objects).}
    \label{fig:lpips-grid}
  \end{subfigure}

  \caption{\textbf{LPIPS-based view selection under the camera sweep protocol.}
  We sweep the elevation from $90^\circ$ (top) to $0^\circ$ (front) in $10^\circ$ steps and vary the camera distance from $1.0\times$ to $2.5\times$ the scene radius (160 renders per scene). 
  While this protocol facilitates a relatively fair and meaningful comparison, the visual bias inherent in a single view allows physically implausible phenomena to be masked under the "best view," leading to overestimated scores. Conversely, our scenes maintain structural integrity and a plausible layout across all perspectives. This indicates that the actual gap between our method and baselines may be larger than the metric score suggests.}
  \label{fig:lpips-sweep}
\end{figure}

\subsection{Visual \& Perceptual Quality Details}

\subsubsection{Camera Sweep Protocol.} Due to the unknown camera perspective of arbitrary input images, we established a standardized camera sweep trajectory for all methods to ensure a fair and meaningful comparison. We generate a total of 160 distinct views per scene. 

The sweep covers the vertical range from a top-down view ($90^\circ$ elevation) to a front view ($0^\circ$ elevation) in $10^\circ$ steps. Concurrently, the camera distance from the scene center is varied in 16 incremental steps, ranging from $1.0\times$ to $2.5\times$ the scene radius. For each generated rendering, we calculate the LPIPS/DINOv2/CLIP scores against the original input image. The reported score of each method on each metric is the best view (the view most closely matching the input image), and the final metric scores are averaged over all scenes.

\begin{table}[tb]
  \caption{\textbf{Best-view and mean-view visual-perceptual comparison with MIDI.} B denotes the best-view protocol used in the main paper, and M denotes the mean-view score averaged over the camera sweep. TabletopGen outperforms MIDI under both protocols, and the relative gain increases under the mean-view protocol.}
  \label{tab:best_mean_view}
  \setlength{\tabcolsep}{3.5pt}
  \centering
  \small
  \begin{tabular}{@{}lcccccc@{}}
    \toprule
    \textbf{Method} & LPIPS-B$\downarrow$ & LPIPS-M$\downarrow$ & DINOv2-B$\uparrow$ & DINOv2-M$\uparrow$ & CLIP-B$\uparrow$ & CLIP-M$\uparrow$ \\
    \midrule
    MIDI       & 0.4559 & 0.5837 & 0.7070 & 0.5027 & 0.8867 & 0.7833 \\
    Ours       & \textbf{0.4483} & \textbf{0.5636} & \textbf{0.8383} & \textbf{0.6500} & \textbf{0.9077} & \textbf{0.8123} \\
    $|\Delta\mathrm{gain}|$ & +0.0076 & \textbf{+0.0201} & +0.1313 & \textbf{+0.1473} & +0.0210 & \textbf{+0.0290} \\
    \bottomrule
  \end{tabular}
\end{table}

\subsubsection{Discussion on Fairness.} Although the best-view selection protocol provides a practical way to compare methods under unknown input camera poses, it can slightly favor certain baselines. As shown in \cref{fig:lpips-grid}, when MIDI generates floating objects or objects exceeding the tabletop boundary, its selected best view ($20^\circ, 1.8\times$) can visually mask these implausible layouts, allowing it to still obtain comparatively good LPIPS/DINOv2/CLIP scores. In contrast, TabletopGen maintains physically grounded layouts across views and is less affected by this bias.

To further examine this effect, we additionally report mean-view results against MIDI, the strongest image-driven baseline in the main paper. The mean-view protocol averages scores over all 160 rendered views instead of selecting only the best-matching one. As shown in \cref{tab:best_mean_view}, TabletopGen remains better than MIDI on all three visual-perceptual metrics, and the gap becomes larger under the mean-view protocol. This supports our observation that best-view matching can underestimate TabletopGen's advantage, since physically invalid baseline layouts may be hidden from a favorable single view.

\subsection{User Study Details}

We conducted a comprehensive human evaluation to compare TabletopGen with baseline methods on tabletop scene quality. In total, 128 unpaid volunteers participated; responses were anonymous and used solely for research.

\begin{figure}[h]
  \centering
    \includegraphics[width=0.85\linewidth]{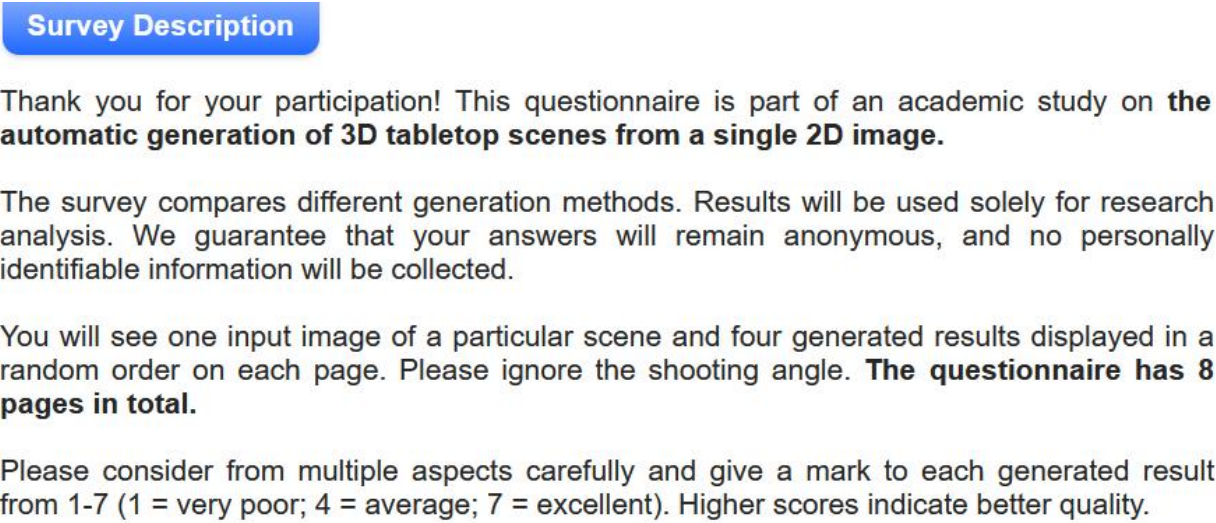}

   \caption{\textbf{Survey description.} Overview of the study goals and instructions.
}
   \label{fig:user_1}
\end{figure}

As shown in \cref{fig:user_1}, the survey began with a brief description of goals and instructions. Each participant was then randomly assigned 8 scenes. For each scene, the interface presented the input image and the four generated results (one per method) in a randomized order to mitigate position bias. To accommodate different audiences, every prompt and criterion was provided in both English and Chinese.

\begin{figure}[t]
  \centering
    \includegraphics[width=0.85\linewidth]{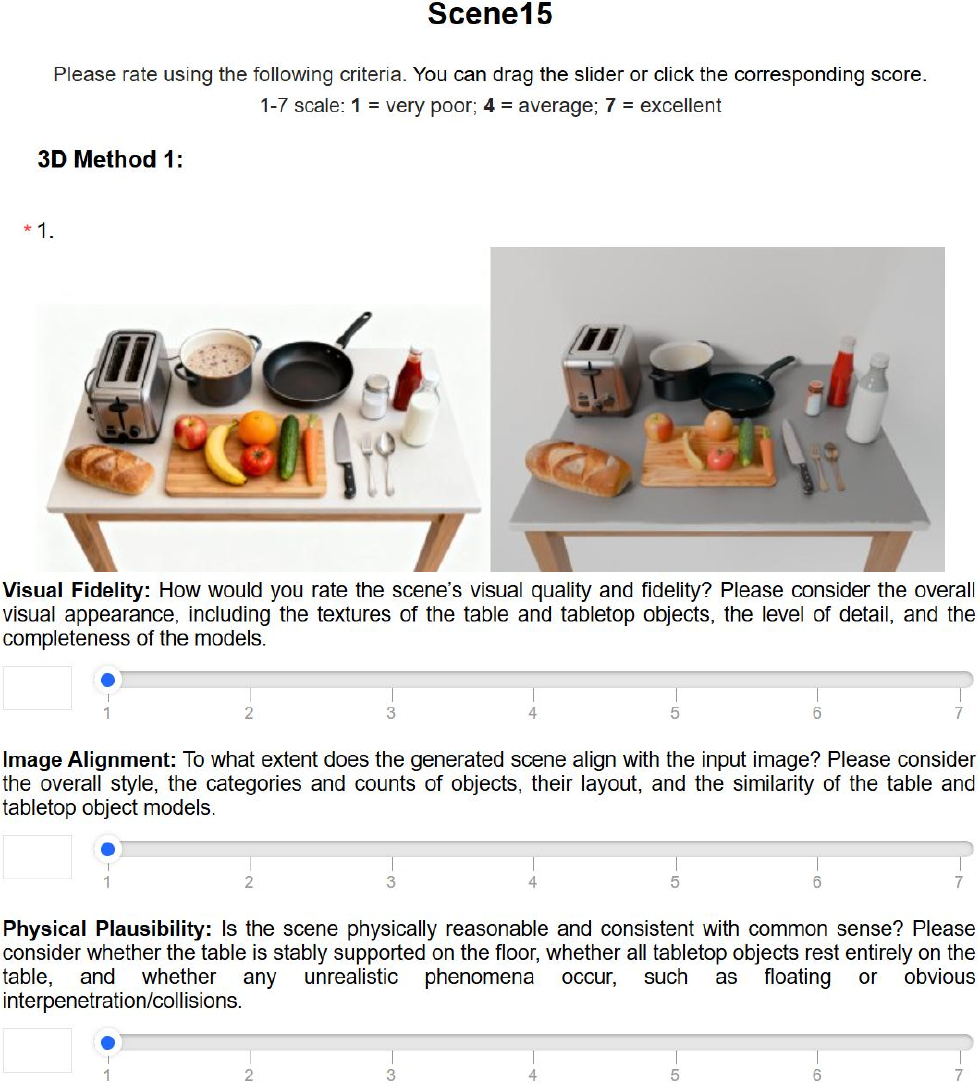}

   \caption{\textbf{Per-scene rating page.} For each scene, participants rate Visual Fidelity, Image Alignment, and Physical Plausibility (1--7) given the input image and one method's result. Shown here is the scoring UI for a single method, the other three methods use the same interface.
}
   \label{fig:user_2}
\end{figure}

For each method, participants rated three criteria on a 1--7 scale (1=very poor, 4=average, 7=excellent) (\cref{fig:user_2}):

\begin{itemize}
    \item \textbf{Visual Fidelity (VF):} overall visual quality and realism of the scene;
    
    \item \textbf{Image Alignment (IA):} consistency with the input image in style, counts, categories, and layout;

    \item \textbf{Physical Plausibility (PP):} physical reasonableness (no floating/penetration, stable support, etc.).
\end{itemize}

\begin{figure}[t]
  \centering
    \includegraphics[width=0.9\linewidth]{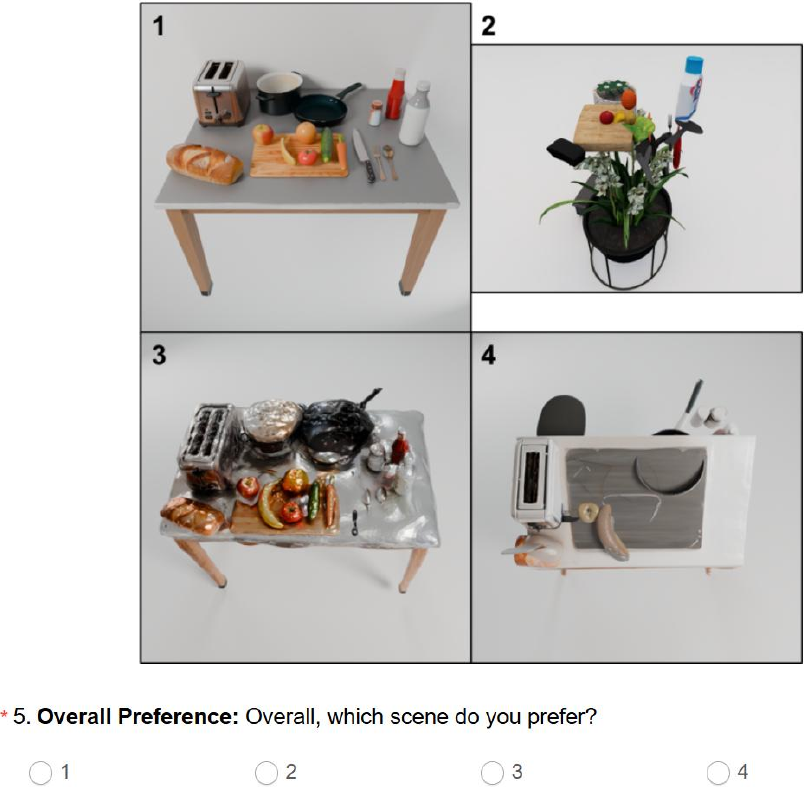}

   \caption{\textbf{Overall preference.} After scoring four methods, participants choose a single overall favorite among the four randomized results.
}
   \label{fig:user_3}
\end{figure}

After scoring the four results, participants selected an Overall Preference (OP) among the four images for that scene (\cref{fig:user_3}).

The mean completion time was 716 seconds. We discarded submissions less than 200 seconds, yielding 126 valid questionnaires. We report per-method means for VF/IA/PP and the OP selection counts and percentages.

\subsection{Prompts}

\begin{figure}[t]
  \centering
    \includegraphics[width=1.0\linewidth]{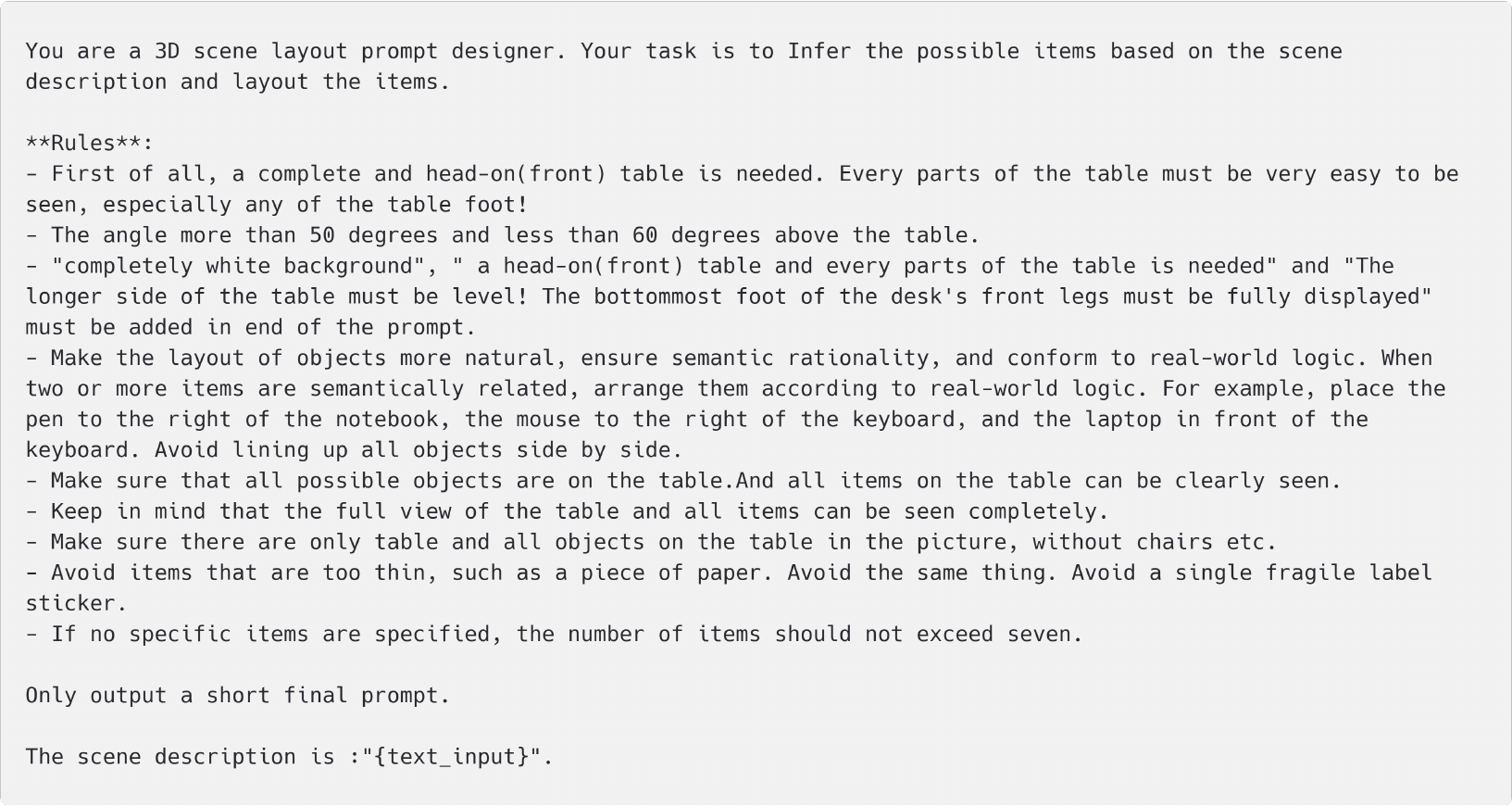}

   \caption{Prompt for expanding a short text input into a detailed scene description.}
   \label{fig:prompt_1}
\end{figure}

\begin{figure}[t]
  \centering
    \includegraphics[width=1.0\linewidth]{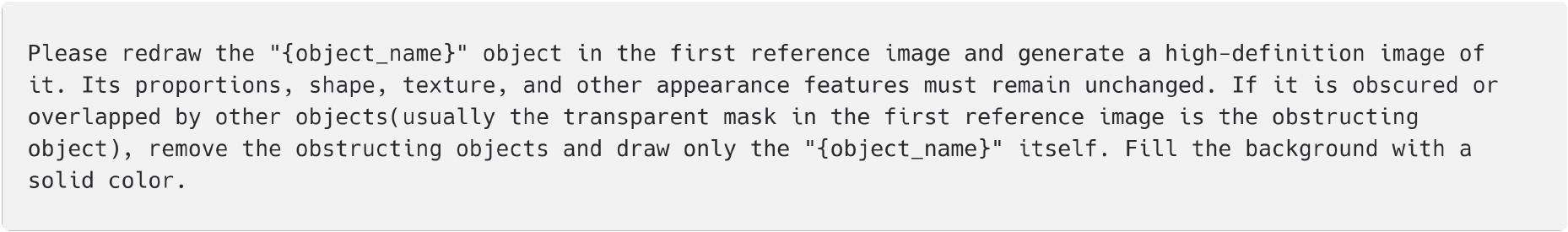}

   \caption{Prompt for multimodal generative completion.}
   \label{fig:prompt_2}
\end{figure}

\begin{figure}[t]
  \centering
    \includegraphics[width=1.0\linewidth]{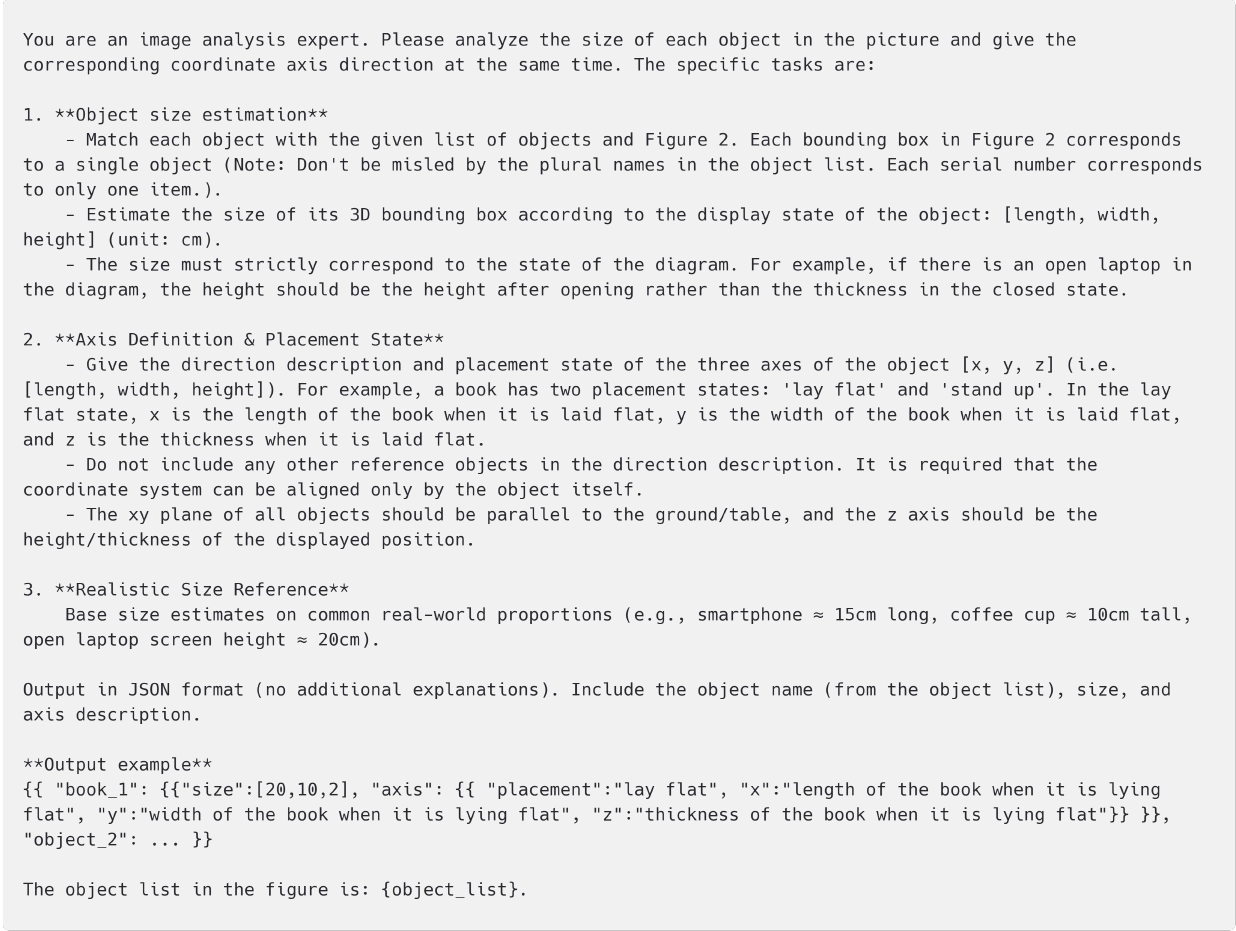}

   \caption{Prompt for analyzing the commonsense size and canonical axis definition of each instance.}
   \label{fig:prompt_3}
\end{figure}

\begin{figure}[t]
  \centering
    \includegraphics[width=1.0\linewidth]{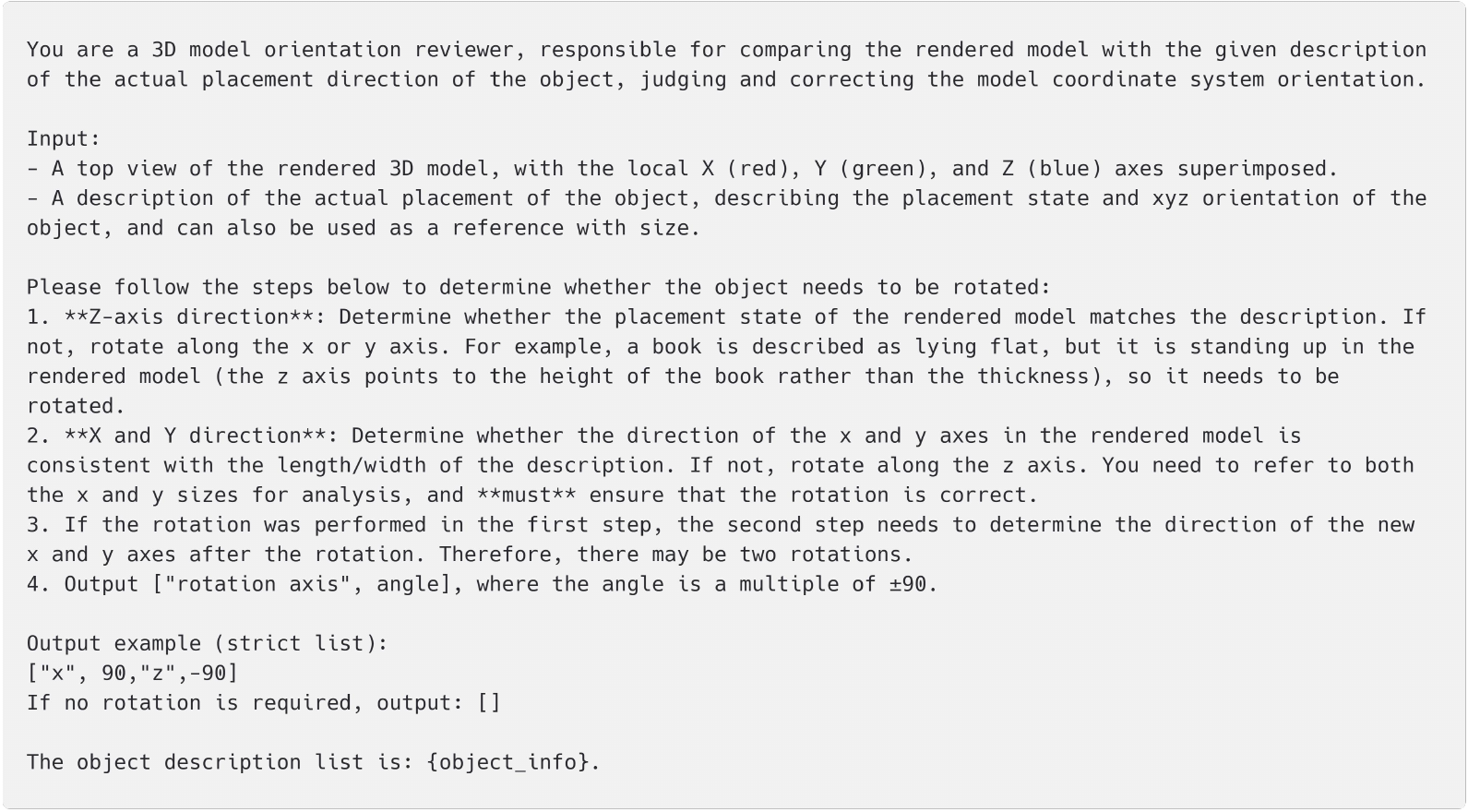}

   \caption{Prompt for canonicalizing the 3D model coordinate system of a single instance.}
   \label{fig:prompt_4}
\end{figure}

\begin{figure}[t]
  \centering
    \includegraphics[width=1.0\linewidth]{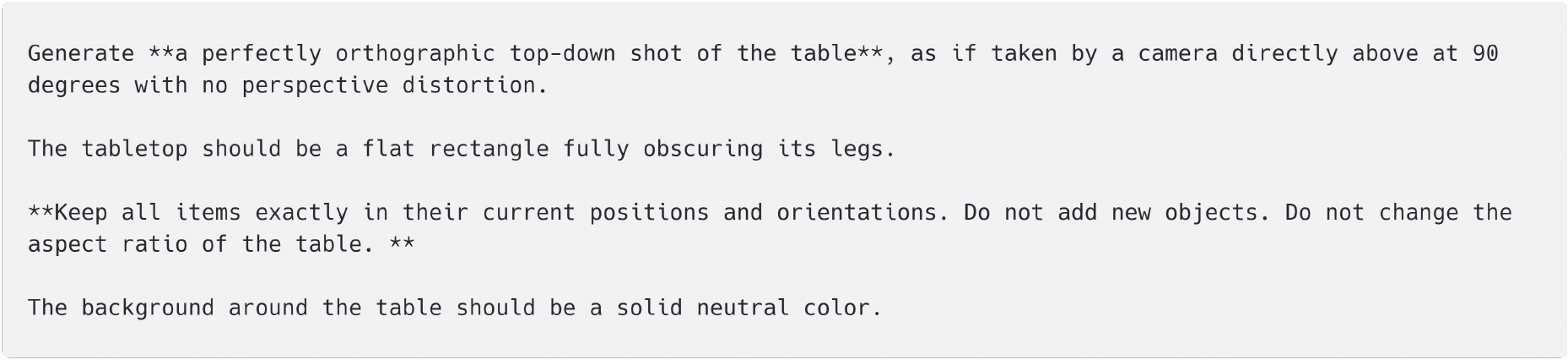}

   \caption{Prompt for synthesizing the top-view image.}
   \label{fig:prompt_5}
\end{figure}

\begin{figure}[t]
  \centering
    \includegraphics[width=1.0\linewidth]{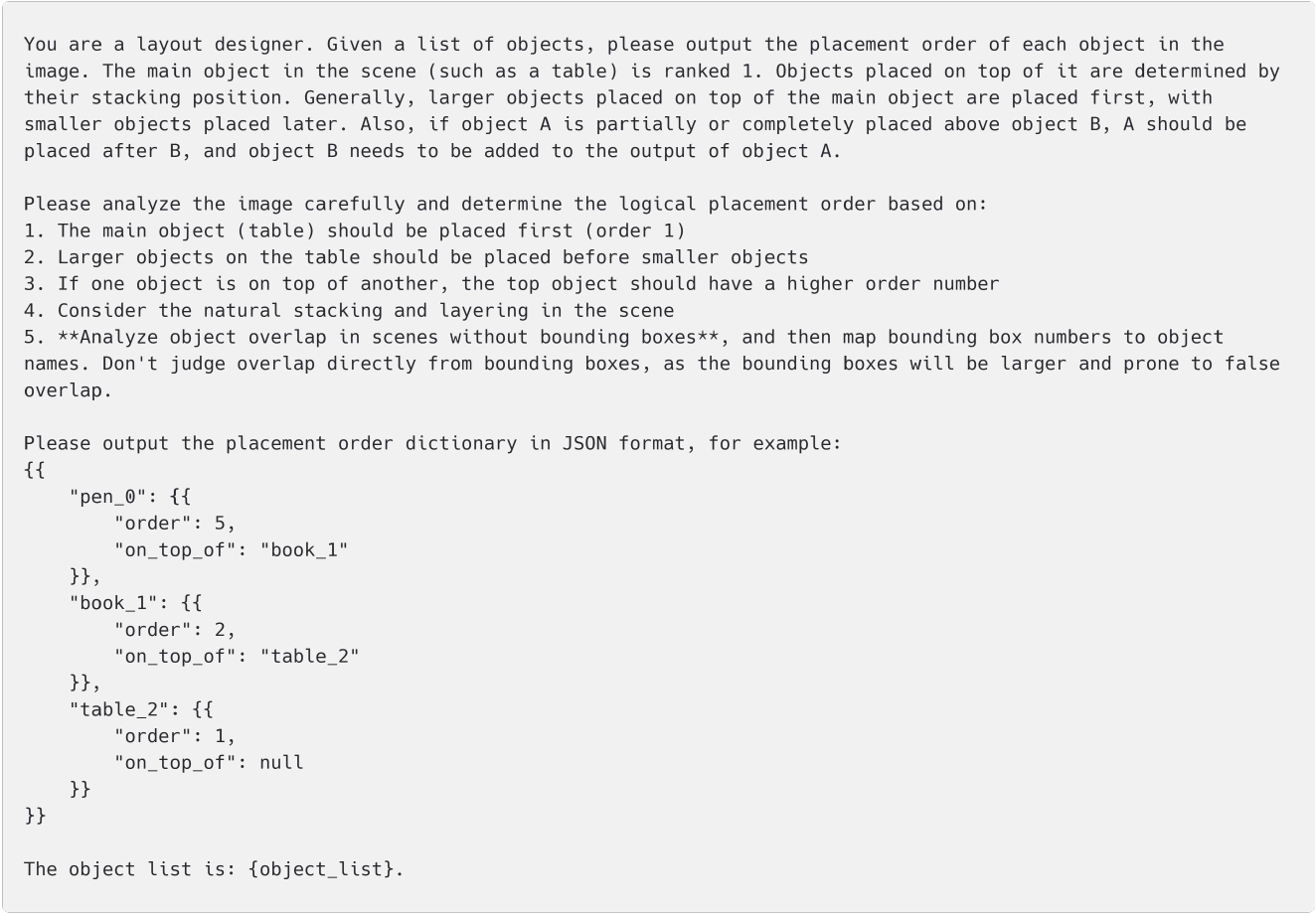}

   \caption{Prompt for analyzing instance stacking order.}
   \label{fig:prompt_6}
\end{figure}

The complete prompt templates used across all stages of TabletopGen are detailed in Figs. \ref{fig:prompt_1}-\ref{fig:prompt_6}.

